\documentclass[lettersize,journal]{IEEEtran}
\usepackage{amsmath,amsfonts}
\usepackage{algorithmic}
\usepackage{algorithm}
\usepackage{array}
\usepackage[caption=false,font=normalsize,labelfont=sf,textfont=sf]{subfig}
\usepackage{textcomp}
\usepackage{stfloats}
\usepackage{url}
\usepackage{verbatim}
\usepackage{graphicx}
\usepackage{cite}
\usepackage{booktabs}
\usepackage{subcaption}
\usepackage{multirow}
\usepackage{threeparttable}
\usepackage{makecell}
\begin{document}

\title{Johnny: Structuring Representation Space to Enhance Machine Abstract Reasoning Ability}

\author{Ruizhuo Song, Member, IEEE,  Beiming Yuan
\thanks{This work was supported by the National Natural Science Foundation of China under Grants 62273036. Corresponding author: Ruizhuo Song, ruizhuosong@ustb.edu.cn}
\thanks{Ruizhuo Song and Beiming Yuan are with the Beijing Engineering Research Center of Industrial Spectrum Imaging, School of Automation and Electrical Engineering, University of Science and Technology Beijing, Beijing 100083, China (Ruizhuo Song email: ruizhuosong@ustb.edu.cn and Beiming Yuan email: d202310354@xs.ustb.edu.cn). }

\thanks{Ruizhuo Song and Beiming Yuan contributed equally to this work.}
}

\markboth{Journal of \LaTeX\ Class Files,~Vol.~14, No.~8, August~2021}%
{Shell \MakeLowercase{\textit{et al.}}: A Sample Article Using IEEEtran.cls for IEEE Journals}


\maketitle

\begin{abstract}

This paper thoroughly investigates the challenges of enhancing AI's abstract reasoning capabilities, with a particular focus on Raven's Progressive Matrices (RPM) tasks involving complex human-like concepts. Firstly, it dissects the empirical reality that traditional end-to-end RPM-solving models heavily rely on option pool configurations, highlighting that this dependency constrains the model's reasoning capabilities. To address this limitation, the paper proposes the Johnny architecture - a novel representation space-based framework for RPM-solving. Through the synergistic operation of its Representation Extraction Module and Reasoning Module, Johnny significantly enhances reasoning performance by supplementing primitive negative option configurations with a learned representation space. Furthermore, to strengthen the model's capacity for capturing positional relationships among local features, the paper introduces the Spin-Transformer network architecture, accompanied by a lightweight Straw Spin-Transformer variant that reduces computational overhead through parameter sharing and attention mechanism optimization. Experimental evaluations demonstrate that both Johnny and Spin-Transformer achieve superior performance on RPM tasks, offering innovative methodologies for advancing AI's abstract reasoning capabilities.
\end{abstract}

\begin{IEEEkeywords}
Abstract Reasoning, Raven's Progressive Matrices, Representation Learning, Structured Representation Space, Inter-Head Communication in Attention Mechanisms.
\end{IEEEkeywords}

\section{Introduction}

\IEEEPARstart {D}{eep} learning, by emulating the mechanisms of the human brain, has demonstrated remarkable performance across numerous domains, particularly in generative tasks such as the creation of images, texts, and videos through technologies like Generative Adversarial Networks (GANs) \cite{GAN}, Variational Autoencoders (VAEs) \cite{VAE}, autoregressive models, and Transformer \cite{Transformer}. Additionally, deep learning has achieved notable success in fields such as computer vision \cite{ResNet, A survey of convolutional neural networks, A survey of visual transformers}, natural language processing \cite{Transformer, GPT-3, Attention in natural language processing}, generative model construction \cite{GAN, VAE, DiffusionModel}, visual question answering systems \cite{VQA, CLEVERdataset}.

Currently, research on Large Language Models (LLMs) within the field of deep learning provides an effective and unified solution framework for the aforementioned domains. However, the academic community has observed that these LLMs \cite{LLM1, LLM2, LLM3} exhibit suboptimal performance in abstract reasoning tasks \cite{LLM1, LLM2, LLM3}. This limitation not only reveals current models' boundaries but also highlights the critical role of abstract reasoning capabilities as an indicator of advanced intelligence. Motivated by these insights, this paper explores methods to enhance the abstract reasoning capabilities of deep learning models.

\section{Problem Description}

In the field of AI, the abstract reasoning ability of mechanical intelligence has drawn considerable academic attention as it is seen as a hallmark of intelligence, essential for AI to attain or surpass human-level intelligence. This capability is vital for AI to operate effectively in complex scenarios, achieve cross-domain knowledge transfer, and generalize knowledge. However, current AI technology, particularly deep learning models, faces limitations in practical applications due to their reliance on large-scale annotated data and struggles with handling abstract concepts, such as those in medical diagnosis and financial analysis, which require causal and counterfactual reasoning. To address this, the academic community has constructed abstract reasoning datasets like RAVEN \cite{RAVENdataset} and PGM \cite{PGMdataset} to quantify AI's reasoning capabilities, but designing algorithms capable of processing symbolic logic, causal relationships, and uncertain reasoning remains a significant challenge. Enhancing abstract reasoning ability requires algorithmic innovation and interdisciplinary theoretical integration, drawing insights from cognitive science and neuroscience. Ultimately, the development of mechanical intelligence's abstract reasoning ability is crucial for achieving artificial general intelligence (AGI), enabling AI to comprehend the world, solve problems, and transfer knowledge flexibly across domains, akin to human intelligence, which will redefine technological boundaries and potentially prompt a reevaluation of human intelligence itself.
This paper focuses on Raven's Progressive Matrices (RPM) problems, delving into and devising effective solutions to enhance the abstract reasoning capabilities of deep learning algorithms.

\subsection{Raven's Progressive Matrices}

The Raven's Progressive Matrices (RPM) problem\cite{RPM} stands as an authoritative benchmark test for evaluating the advanced discriminative capabilities of deep learning algorithms, encompassing abilities such as abstract reasoning, pattern recognition, and problem-solving. As a sophisticated non-verbal task, RPM demands that algorithms possess the capacity for knowledge generalization and transfer, thereby incentivizing them to exhibit cognitive functions that approximate human intelligence. Its standardized nature and strong correlation with human intelligence metrics render RPM a pertinent and consistent test for assessing the complexity of machine learning models, further underscoring its authority in gauging the intelligence level of deep learning algorithms. Currently, the RAVEN dataset \cite{RAVENdataset} and the PGM dataset \cite{RAVENdataset} are two widely recognized instances of RPM problems.

\begin{figure}[htp]\centering
	\includegraphics[trim=11cm 0cm 0cm 0cm, clip, width=7.5
 cm]{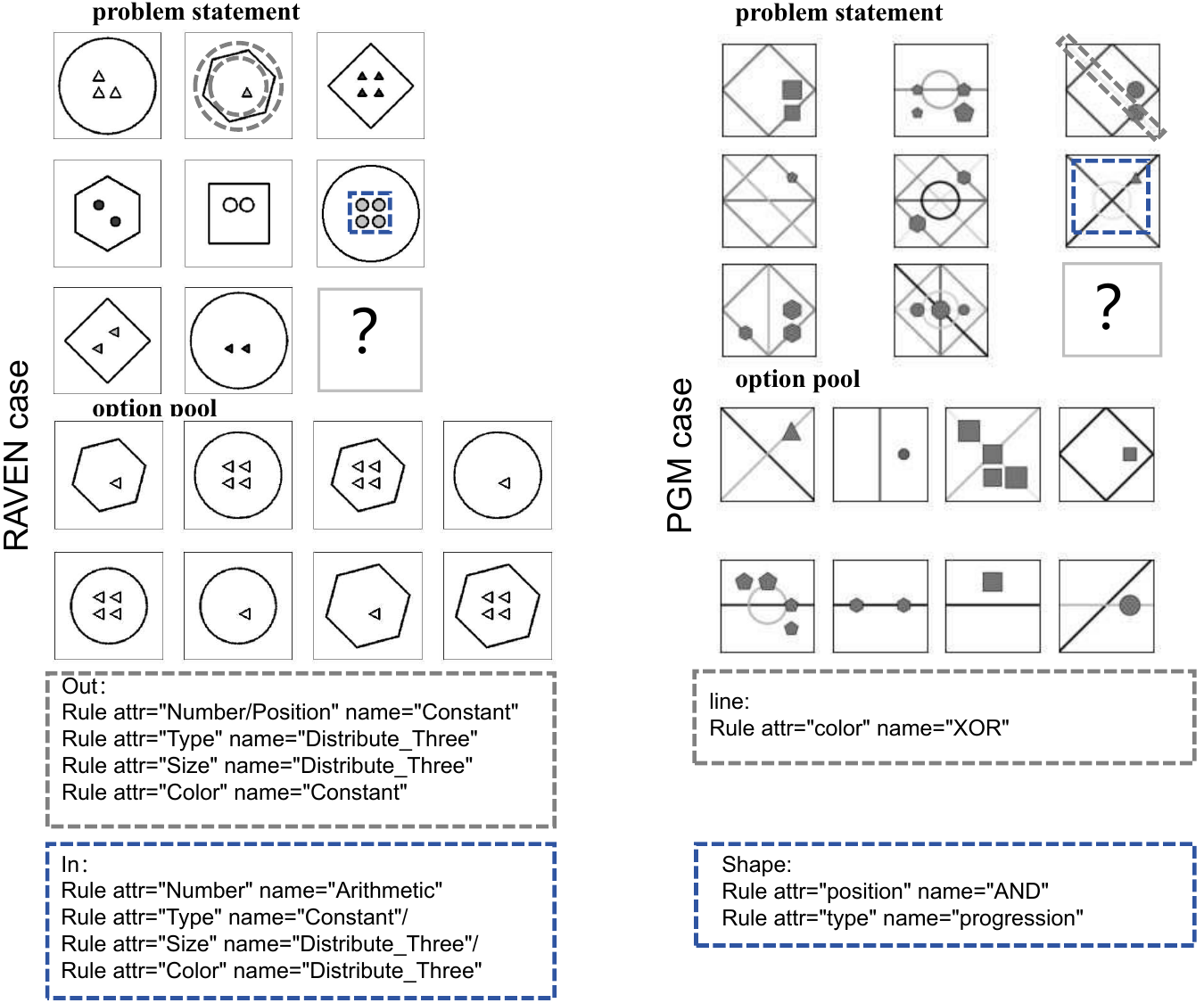}
	\caption{RAVEN and PGM case}
\label{RAVEN}
\end{figure}

The RAVEN database is a collection of Ravens Progressive Matrices instance, each consisting of 16 images. Eight of these images, known as the ``statement," are the main challenge, while the other eight form a pool for option. The goal is to choose the right images to complete a $3\times3$ matrix with a progressive geometric pattern, following an abstract ``rule." The left side of Figure \ref{RAVEN} shows an example of RAVEN. Similarly, PGM provides eight images for the statement and options, but its ``rule" applies to both rows and columns. The right side of Figure \ref{RAVEN} gives a typical example of PGM.

\subsection{Related work}

Image reasoning models such as CoPINet\cite{CoPINet}, LEN+teacher\cite{LEN}, and DCNet\cite{DCNet} focus on learning disparities and rules. Meanwhile, NCD\cite{NCD}, SCL\cite{SCL}, SAVIR-T\cite{SAVIR-T}, and neural symbolism systems (PrAE, NVSA, ALANS\cite{PrAE,ALANS,NVSA}) enhance both interpretability and accuracy. RS-CNN and RS-TRAN\cite{RS} excel in solving RPM problems, while Triple-CFN\cite{Triple-CFN} stands out by implicitly extracting and indexing concept and reasoning unit info, boosting reasoning accuracy. CRAB\cite{CRAB}, based on Bayesian modeling, has made strides with a tailored conservatory for the RAVEN database, though core challenges remain untackled, leaving the scientific community eager for the broader impact of these innovative methods.

\section{Methodology}

This paper analyzes the general formulation of Raven's Progressive Matrices (RPM) solving models and discovers that the limitation in the reasoning capability of these solving models stems from the insufficiency of option configurations in RPM problems. Based on this finding, the paper proposes the necessity of constructing a representation space for RPM-solving models and utilizing this space to supplement the original option configurations in RPM problems, thereby assisting the solving models in comprehending a more complete problem space.

\begin{figure}[ht]\centering
	\includegraphics[trim=0cm 0cm 0cm 0cm, clip, width=8cm]{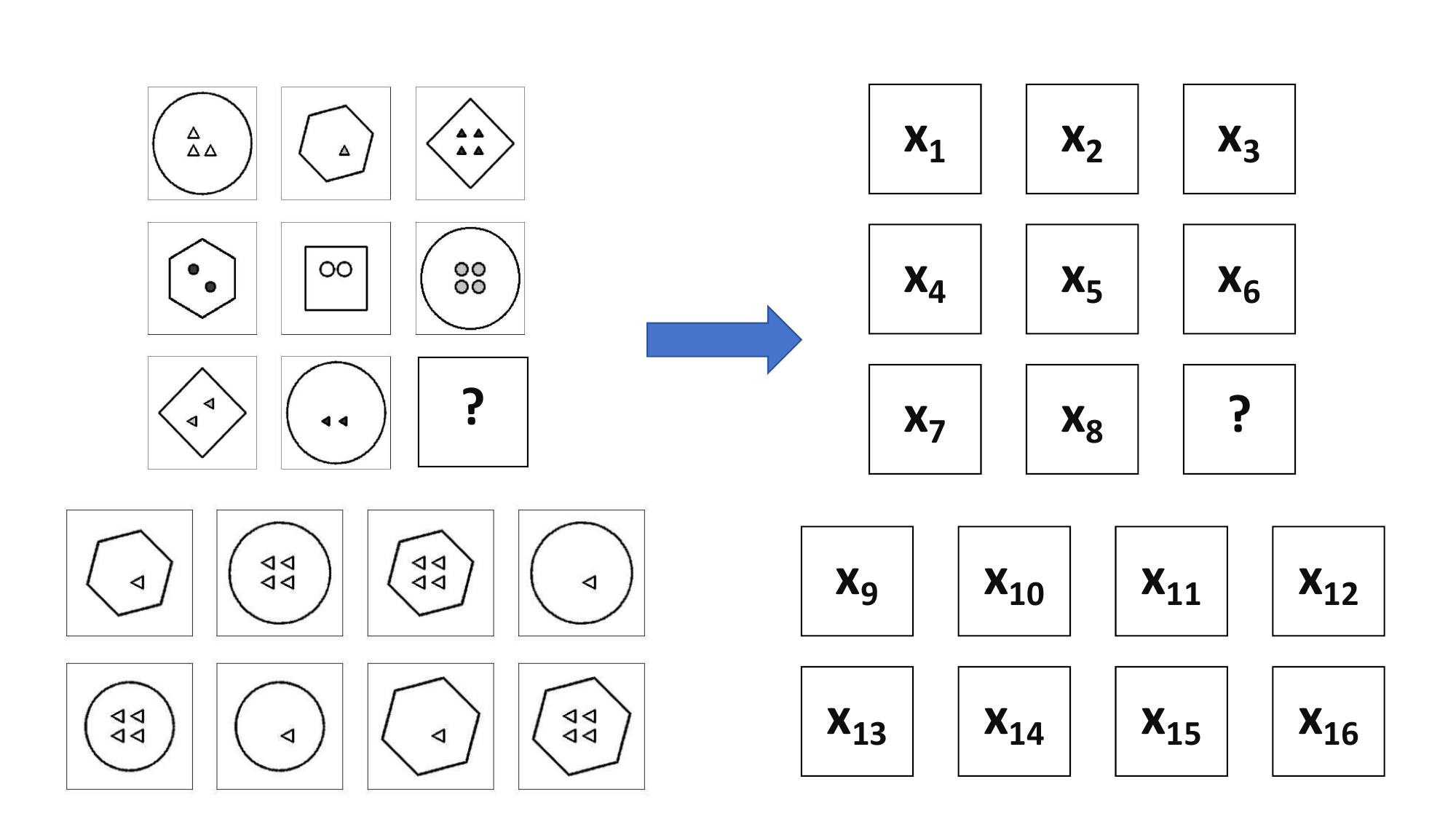}
	\caption{ Annotations of images within a RPM instance}
\label{annotates}
\end{figure}

This paper annotates the images within a specific RPM instance as shown in the Figure \ref{annotates}.

\section{ACT1: The drawbacks of end-to-end RPM-solving models.}

In this section, we have analyzed the general formulation of end-to-end RPM-solving models and clarified the drawbacks of this formulation.

\subsection{The General Form of End-to-end RPM-solving Models.}

\begin{figure}[ht]\centering
	\includegraphics[trim=0cm 3cm 5cm 0cm, clip, width=8cm]{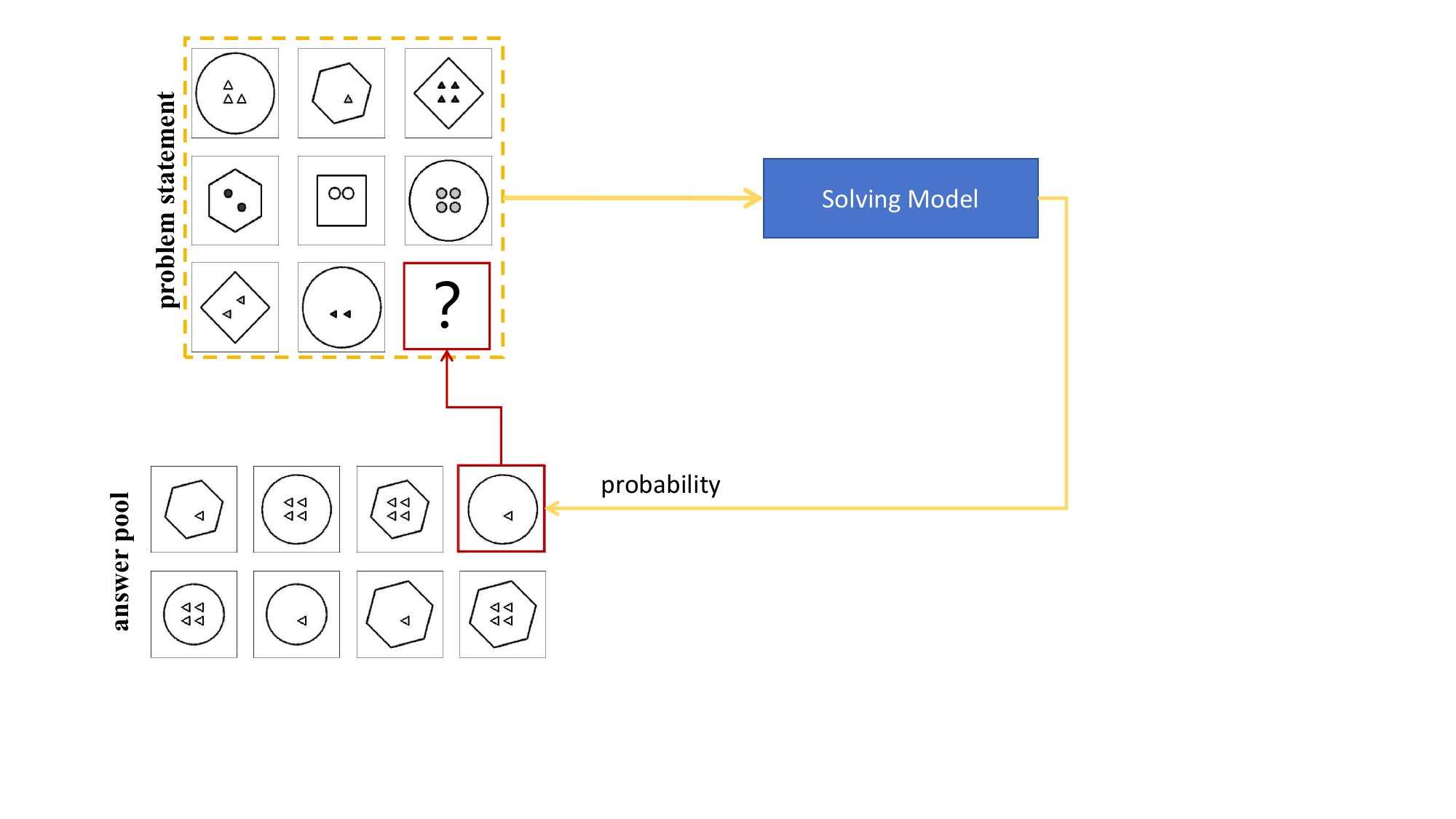}
	\caption{The diagram of the structure of end-to-end RPM-solving models}
\label{end-to-end-frame-work}
\end{figure}

For RPM problems, if the completed matrix adheres to the progressive patterns derivable from the problem statement, the corresponding option is identified as correct \cite{RPMInductivebias}. Building on this observation, most high-performance end-to-end RPM-solving models \cite{CoPINet,SCL,SAVIR-T,Triple-CFN,RS} adopt the architecture depicted in Figure \ref{end-to-end-frame-work}. These models primarily aim to compute, for each candidate in the option pool of an RPM instance, a probability value that quantifies its likelihood of being the correct answer.

Undoubtedly, under the condition of lacking supervisory signals other than correct option indices, this model demonstrates extremely significant and effective advantages.

\subsection{A Fresh Perspective on End-to-end RPM-solving Model}
We might as well regard the end-to-end RPM-solving model with the aforementioned pattern as a probability distribution. Matrices completed with correct options are highly likely to conform to this distribution, while matrices completed with incorrect options are less likely to do so. Training the RPM-solving model can be viewed as optimizing and adjusting this probability distribution.

\subsection{The Training Process of End-to-end RPM-solving Models.}

This paper proposes that the training of end-to-end RPM-solving models can be viewed as the optimization and adjustment of a probability distribution. If we assume that the form of this probability distribution is Gaussian, then optimizing this distribution can be regarded as adjusting the mean and variance of the distribution such that the correct option is assigned a high probability and the incorrect options are assigned low probabilities. Specifically, we adjust the parameters $\mu_\theta$ and $\sigma_\theta^2$ in Formula \ref{P-RPMSoLvers} so that substituting the correct option into $x_\alpha$ in Formula \ref{P-RPMSoLvers} yields the minimum value, while substituting the incorrect options into $x_\alpha$ results in the maximum value.
\begin{align}\label{P-RPMSoLvers}
    -\log(P(x_\alpha)) = \frac{{(x_{a} - \mu_\theta)^{2}}}{{2\sigma^{2}_\theta}} - \log(|\sigma_\theta|)-\frac{1}{2}\log(2\pi)
\end{align}

It is apparent that when the correct option is plugged into $x_{\alpha}$, the Euclidean distance between $x_{\alpha}$ and $\mu_\theta$ should be \textbf{minimized} to obtain the minimum value of Formula (\ref{P-RPMSoLvers}); conversely, when an incorrect option is substituted for $x_{\alpha}$, $\sigma_\theta^{2}$ should be \textbf{minimized} to obtain the maximum value of Formula (\ref{P-RPMSoLvers}). Since $\mu_\theta$ represents the center of the Gaussian distribution and $\sigma_\theta^{2}$ describes the spread of the Gaussian distribution, we can deduce that during the training of the RPM-solving model, it converges the boundary based on the observed incorrect options and determines the center based on the observed correct options.

\subsection{The Drawback of End-to-end RPM-solving Models.}
The aforementioned training process indicates that the end-to-end RPM-solving model heavily relies on observed incorrect options. The extent to which incorrect options assist the solving model in converging the boundary determines the reasoning accuracy of the solving model. This implies that, theoretically, the solving model is expected to achieve optimal performance only when all possible forms of incorrect options are enumerated for the end-to-end RPM-solving model.

Although the form of the RPM-solving model mentioned in Figure \ref{end-to-end-frame-work} has demonstrated undeniable superiority and effectiveness under conditions with limited supervisory signals, and even most large models learning visual abstract reasoning problems adhere to this form, its reliance on the comprehensiveness of the provided incorrect options cannot be overlooked.

\section{ACT2: The completion of incorrect options releases the performance of the RPM-solving model}

In the previous section, we analyzed the paradigm and training process of the end-to-end RPM-solving model, revealing it converges boundaries based on observed incorrect options. Theoretically, to maximize the model's potential, all possible incorrect option forms must be enumerated during training; otherwise, it may suffer from loose boundary convergence and suboptimal performance.

\subsection{Feasibility of Enumerating Option Configuration} Is it feasible to directly adopt the approach of enumerating all the incorrect options when training an end-to-end RPM-solving model, with the aim of obtaining a model that possesses strong reasoning intelligence?
The answer is no, and the reasons are as follows:

\begin{enumerate}
    \item Although in theory this could significantly enhance the performance of the solving model, this training method that leans towards an expert system lacks intelligence.
    \item In RPM, certain reasoning attributes are continuous and cannot be enumerated. Even if, in some specific RPM instances, the possible values of attributes can be enumerated, which allows for the widespread application of enumerating method, this approach does little to contribute to the development of general methods for enhancing AI's abstract reasoning capabilities.
    \item A methodology that can unlock exceptional performance from incomplete option configurations and limited supervisory signals represents a highly effective approach for advancing AI's reasoning capabilities. It is precisely the kind of endeavor that deep learning research should wholeheartedly commit itself to pursuing.
\end{enumerate}
Therefore, this paper does not intend to forgo the opportunity to improve the performance of the solving model by enhancing the configuration of incorrect options. 

\subsection{Sub-enumeration Technology}

This paper proposes that the representations of observable images in RPM problems—particularly those used for training—can be extracted via deep learning models, and the spatial relationships among these representations can be modeled as a bounded, discrete space. By leveraging the structural components of this modeled space, we refine the option configuration of the solving model, thereby developing a sub-enumeration technique. This framework enables an enumeration technique for incorrect option configurations based on the representation space, thereby facilitating the solving model to approach its convergence boundary.

\begin{figure}[ht]\centering
	\includegraphics[trim=0cm 0cm 0cm 0cm, clip, width=8cm]{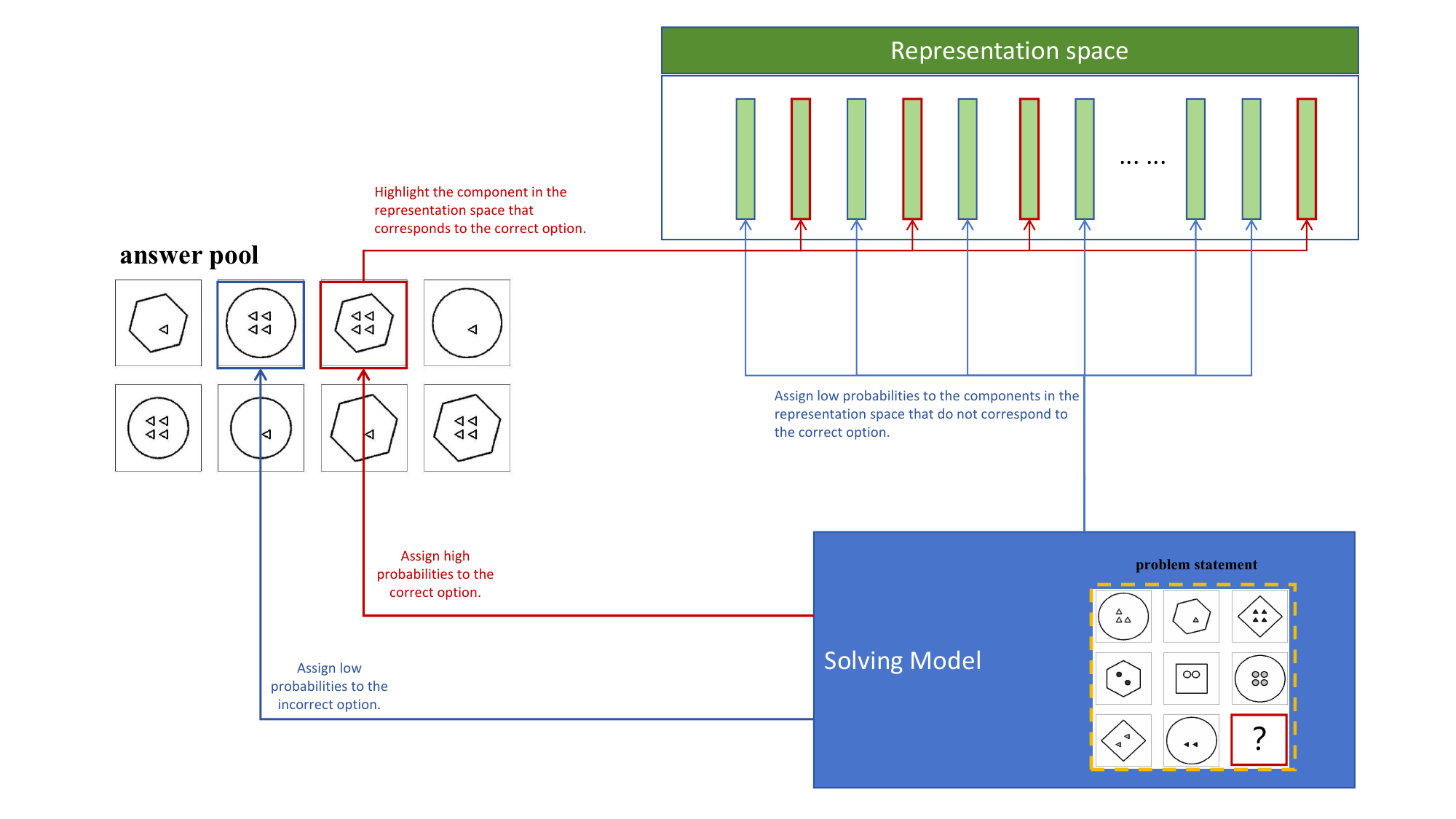}
	\caption{RPM-solving model training process augmented with representation space.}
\label{training process augmented with representation space.}
\end{figure}

Specifically, for all observable RPM images, we extract their representations using deep learning algorithms and construct a representation space. This space comprises discrete components whose flexible combinations can effectively represent the representations of any observed RPM images. Once constructed, this space can be utilized to augment the training process of RPM-solving models. During the model's regular training, we incorporate an additional learning task based on this representation space.
This implies that the RPM-solving model must not only correctly identify the representations of correct options (assigning high probabilities) and exclude incorrect options (assigning low probabilities), but also distribute appropriate probabilities across components within the representation space to reflect their correctness, particularly ensuring that components not corresponding to correct options are also assigned low probabilities. The entire training process is illustrated in Figure \ref{training process augmented with representation space.}.

Our objective is unambiguous: to construct a bounded, discrete representation space comprising a finite set of components. The components within this representation space are capable of encoding all observable RPM images. When the RPM-solving model comprehensively learns these components, it enables the model to clearly perceive the comprehensive configurations of incorrect options. This implements an effective sub-enumeration technique as an alternative approach to enumerating incorrect option configurations.

\section{ACT3: Construction of RPM Image Representation Space and Compatible RPM-solving Model}\label{ACT3}

It seems that the architecture of an RPM-solving model capable of being compatible with representation spaces has already been determined, as this solving model has the following requirements:
\begin{enumerate}
    \item Since this RPM-solving model must be paired with an explicit representation space, it requires a module for extracting image representations and a reasoning module capable of operating on RPM image representations. In other words, equipping the RPM-solving model with a representation space requires explicitly separating the representation extraction process from the reasoning process.
    \item Requirements of Representation Extraction Module: Given that the representations of RPM images observable by the solving model must be expressible within a representation space with a finite and countable number of components, the representations extracted by the solving model for RPM images are necessarily tokenized.
    \item Requirements of Reasoning Module: When the reasoning module processes tokenized RPM image representations, it should adopt the approach of independently evaluating each representation token. In other words, it cannot assume non-independence or dependency among the tokens in the RPM image during reasoning. Only a reasoning module of this form can evaluate all representation tokens of the options while simultaneously assessing the entirety of components within the representation space. Otherwise, the method for evaluating the representation space would have an extremely high order of magnitude.
\end{enumerate}

\begin{figure}[ht]\centering
	\includegraphics[trim=0cm 0cm 0cm 0cm, clip, width=8cm]{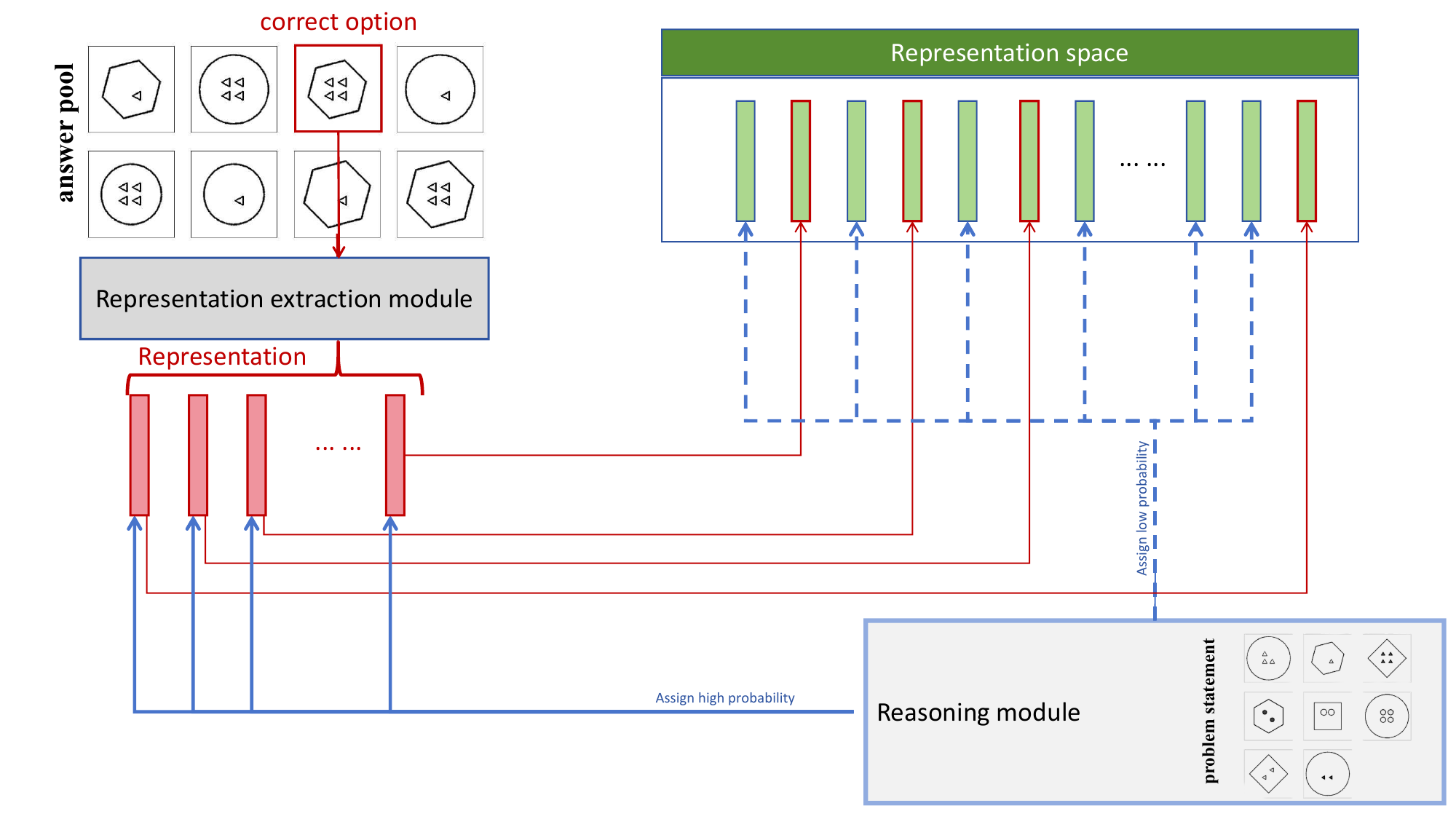}
	\caption{RPM-solving model capable of being compatible with representation spaces}
\label{RPM-solving model capable of being compatible with representation spaces}
\end{figure}

Taking into account the architectural requirements of the RPM-solving model mentioned above, this architecture can be constructed by referring to the approach shown in Figure \ref{RPM-solving model capable of being compatible with representation spaces}.
In this paper, we refer to this structure as Johnny (Joint-Representation-space-based High-Performance Neural Network Yielding Paradigm).

\subsection{Design of the Representation Extraction Module.}
The representation extraction process capable of tokenization can ensure the existence of a representation space with a countable number of components. Currently, the options for models that can achieve tokenized image representation extraction are relatively limited. In this paper, we choose the Visual Transformer (ViT)\cite{ViT} as the backbone for the representation extraction module of Johnny.

\begin{figure}[ht]\centering
	\includegraphics[trim=0cm 0cm 0cm 0cm, clip, width=8.5cm]{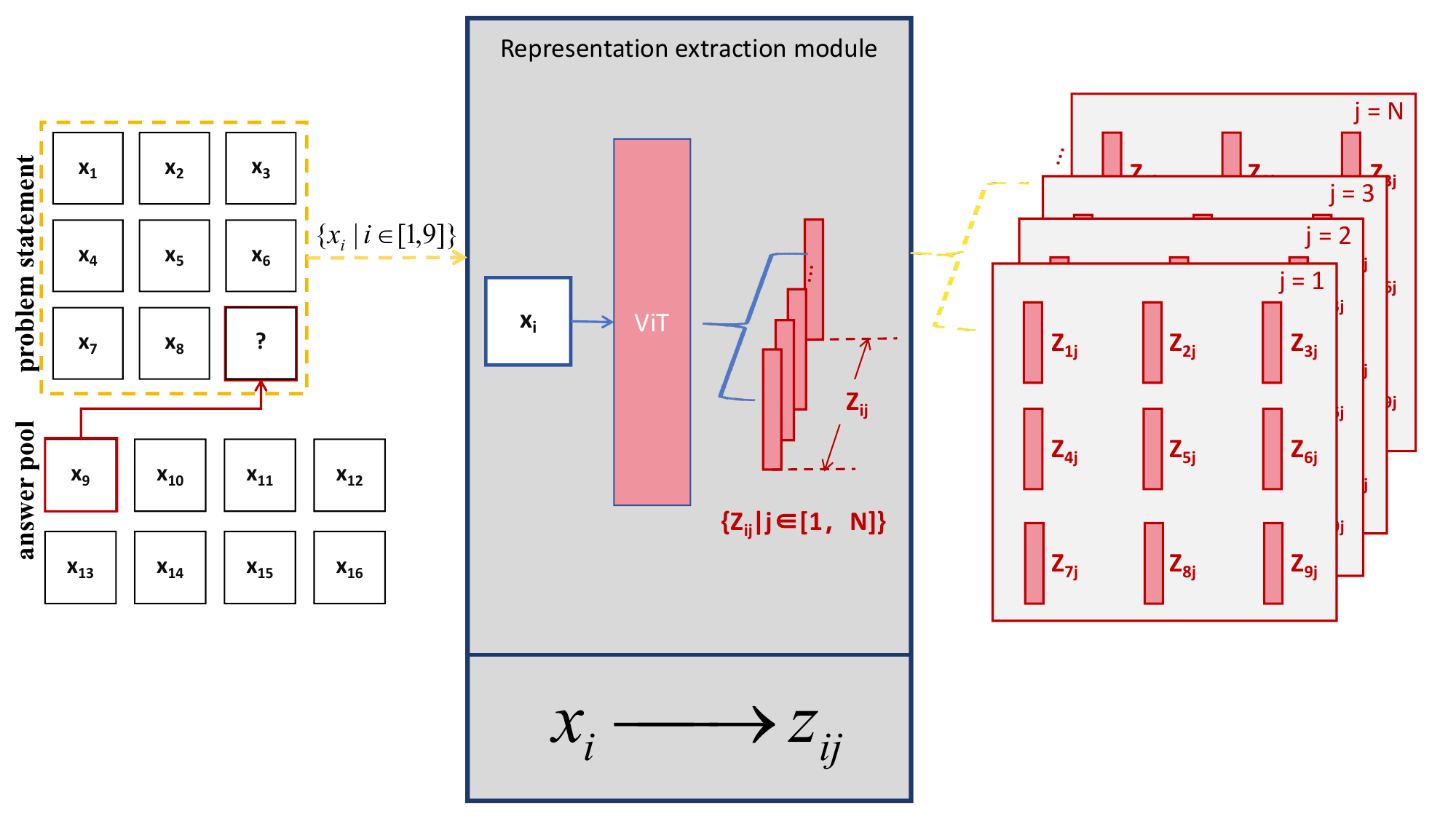}
	\caption{The process by which Johnny's representation extractor processes RPM images}
\label{process of representation extractor}
\end{figure}

Figure \ref{process of representation extractor} illustrates the process by which Johnny's Representation Extraction Module processes RPM images, that is, how the ViT processes the RPM image matrix $\{x_i| i \in [1,9]\}$ into a multi-token representation matrix $\{z_{ij} | i \in [1,9], j \in [1,N]\}$.
In the figure \ref{process of representation extractor}, $N$ represents the number of tokens generated when ViT transforms a single RPM image into a tokenized representation, which is equivalent to the number of image patches divided during the image splitting process in ViT.

\subsection{Design of the Reasoning Module.}
\begin{figure}[ht]\centering
	\includegraphics[trim=0cm 0cm 0cm 0cm, clip, width=8.5cm]{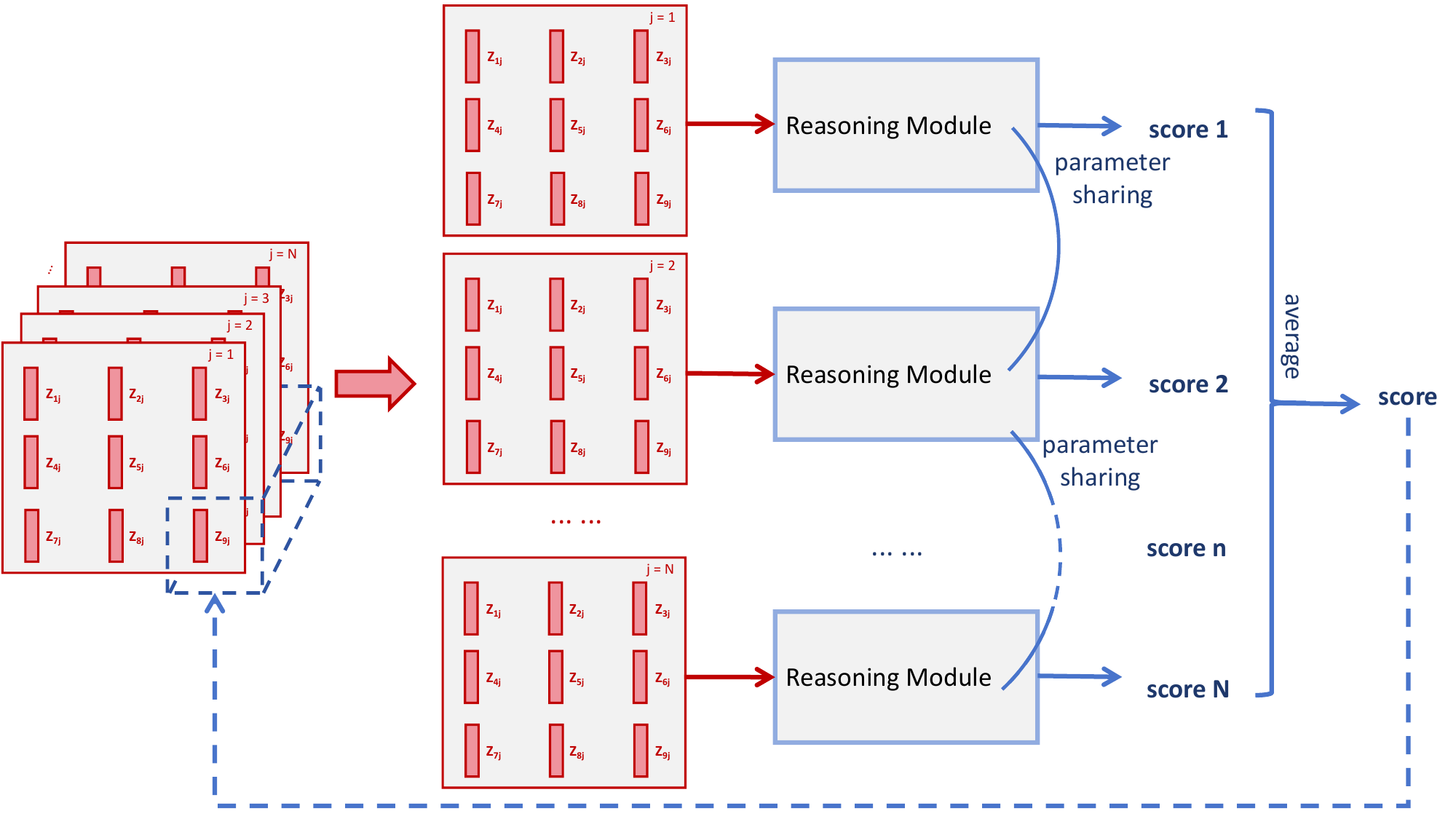}
	\caption{The process by which Johnny's Reasoning Module processes RPM representations}
\label{The process by which Johnny's Reasoning Module processes RPM representations}
\end{figure}

Johnny's Reasoning Module independently processes the $N$ tokens extracted from the RPM image by the Representation Extraction Module. This means that for each RPM option, the Reasoning Module independently performs $N$ evaluations, and finally determines the option's score by calculating the average of these $N$ evaluation results, which intuitively reflects the probability of the option being the correct answer. 
This process can be presented as Figure \ref{The process by which Johnny's Reasoning Module processes RPM representations}.

\begin{figure}[ht]\centering
	\includegraphics[trim=0cm 0cm 0cm 0cm, clip, width=8.5cm]{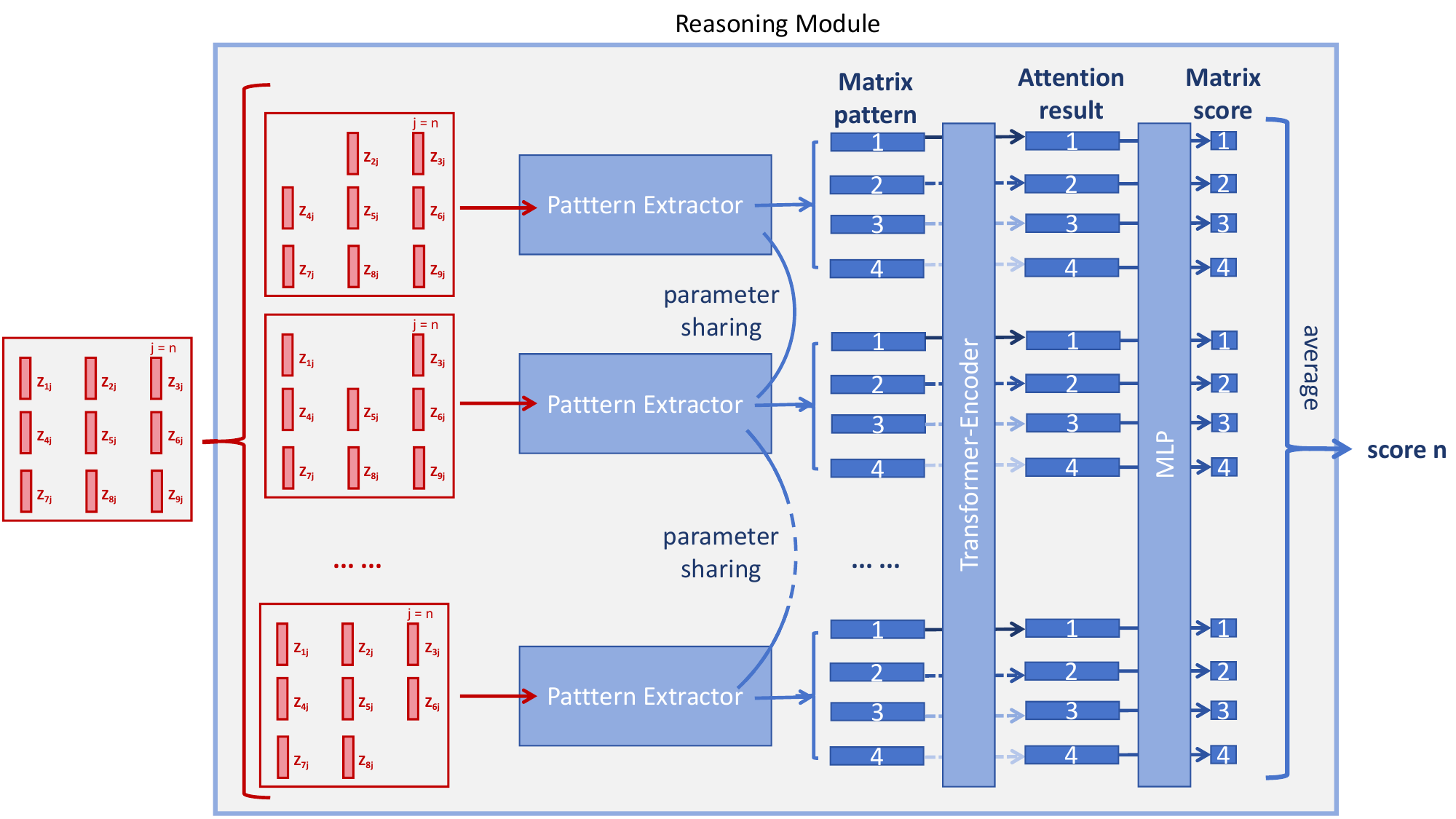}
	\caption{The architecture of the Reasoning Module.}
\label{The architecture of the reasoning module.}
\end{figure}

The architecture of the reasoning module are shown in the Figure \ref{The architecture of the reasoning module.}. 
\begin{enumerate}
    \item As indicated by the red line in the figure, the Reasoning Module first performs an item-wise deduction operation on a single representation matrix. For a $3\times 3$ matrix containing 9 independent element items, the systematic processing yields nine distinct combinations of deduction results.

    \item The figure also shows that the nine deduction results are processed by the Pattern Extractor inside the Reasoning Module into a vector group composed of four vectors, which is recorded as the matrix pattern.

    \item Subsequently, the equivalent vectors in the nine vector groups are processed by the Reasoning Module using Transformer-Encoder for dimension-equivariant processing. The resulting attention results are processed by an MLP with a bottleneck structure into a one-dimensional scalar, which represents a logic score. Finally, the total logic score for the $n$-th token (score $n$) in the image representation is obtained by averaging all the individual logic scores.
\end{enumerate}

\begin{figure}[ht]\centering
	\includegraphics[trim=0cm 0cm 0cm 0cm, clip, width=8.5cm]{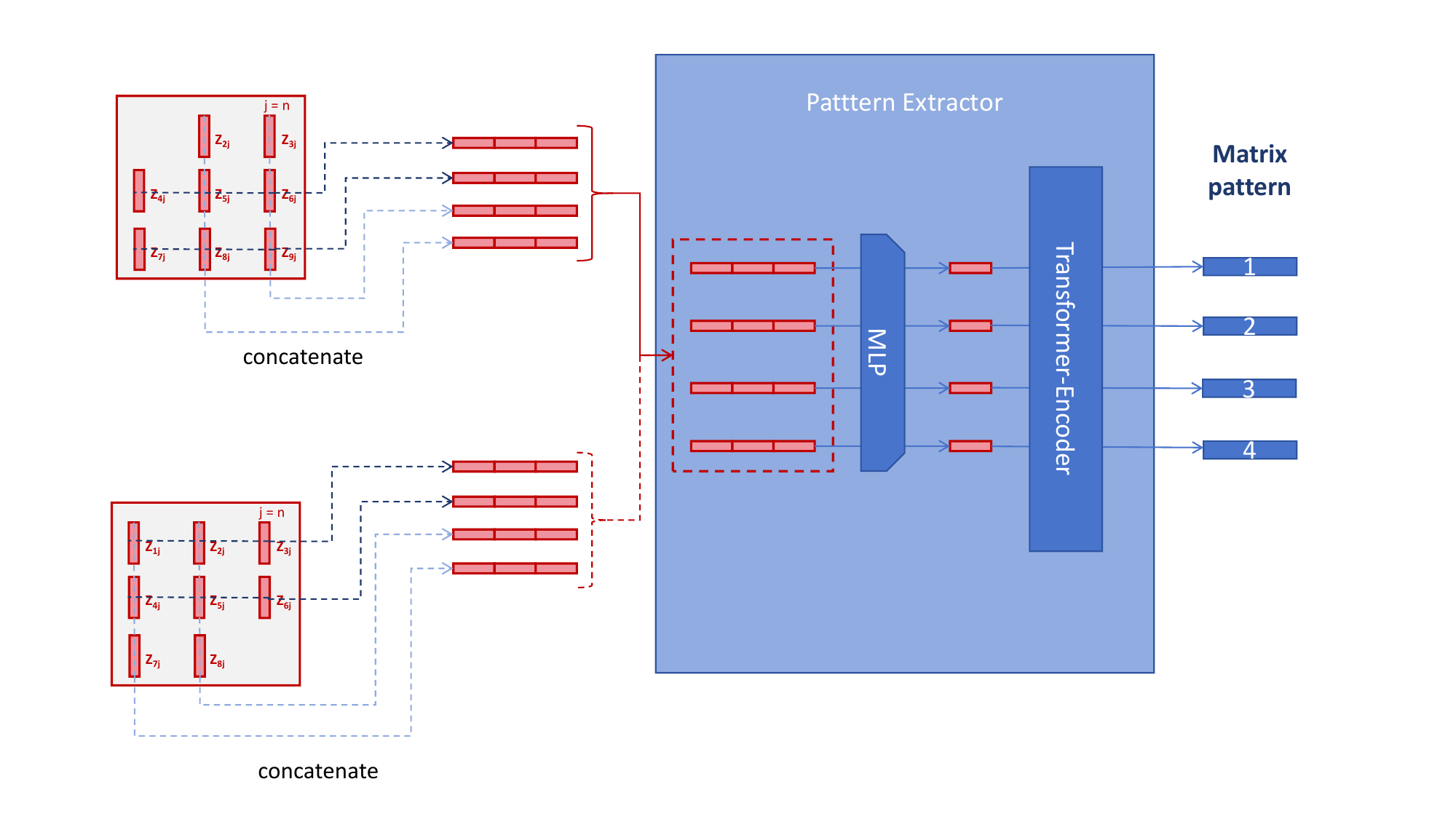}
	\caption{The architecture of the Pattern Extractor.}
\label{The architecture of the Pattern Extractor.}
\end{figure}

As for the Pattern Extractor in the Reasoning Module, its design is relatively straightforward, and its structure is illustrated in the Figure \ref{The architecture of the Pattern Extractor.}.
The figure illustrates the feedforward process of the Pattern Extractor, taking two types of deducted item representation matrices as examples. 
\begin{enumerate}
    \item As depicted, the Pattern Extractor first concatenates the representations within each deducted item matrix along both the row and column directions. 
    
    \item After this concatenation step, a multi-layer perceptron (MLP) with a bottleneck structure is used to fuse the concatenated representations, with the aim of extracting progressive information across the row and column directions.

    \item Subsequently, these four pieces of progressive information are fed into the Transformer-Encoder.

    \item Finally, the logical relationships extracted by the Transformer-Encoder are regarded as the inherent matrix pattern embedded within the original item representation matrix.
\end{enumerate}

After elucidating the architecture and functionality of the Representation Extraction Module and the Reasoning Module, Johnny has already developed fundamental reasoning capabilities. During the training of Johnny, who is equipped solely with these two modules, this paper utilizes the Cross-Entropy loss function, adhering to the logical framework for problem-solving, to impose constraints on the scores that Johnny calculates for each available option.

\subsection{Design of the Representation Space.}
The structural of the representation space is not particularly distinctive. Essentially, the representation space is a set of vectors, where the components within it are optimizable and learnable. We define this representation space as $\{T_k | \, k \in [1, K]\}$. Here, $K$ denotes the number of components (i.e., vectors) in the space. And $T_k$ is an optimizable vector whose dimension is consistent with that of a single token in the representation extracted by the Representation Extraction Module from the RPM image.

\begin{figure}[ht]\centering
	\includegraphics[trim=0cm 0cm 0cm 0cm, clip, width=8.5cm]{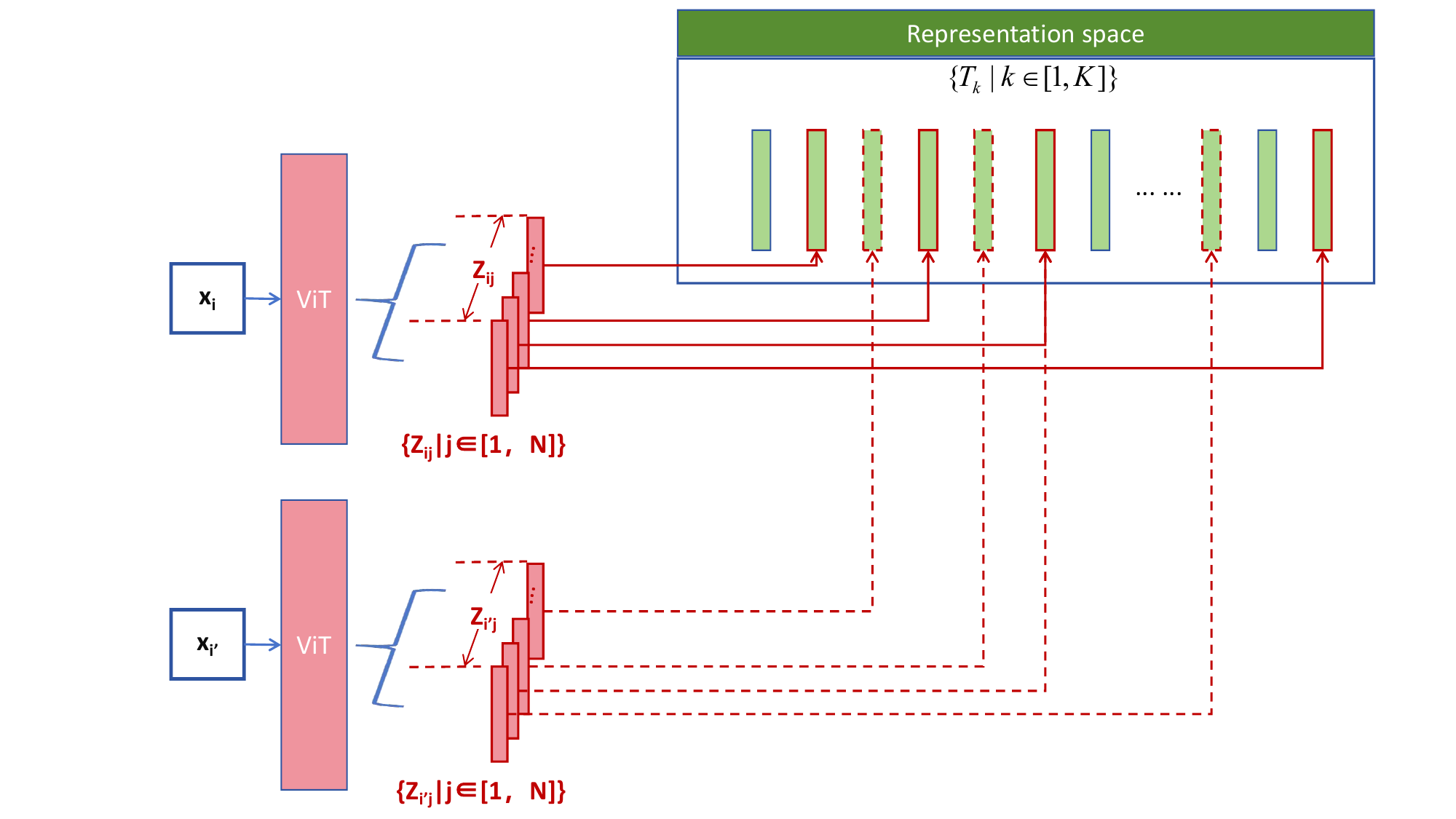}
	\caption{The architecture of the Representation Space.}
\label{The architecture of the Representation Space.}
\end{figure}

This paper puts forward the requirement that all RPM images observed by Johnny's Representation Extraction Module should be able to be represented by the components in the representation space. In other words, we hope that every token in the representation of any RPM image used for training can find a corresponding component within this representation space. Any $z_{ij}$ can find its corresponding counterpart $T_{\tilde k}$ ($\tilde k \in [1,K]$) within the set $\{T_k | k \in [1, K]\}$. Johnny needs to establish an alignment between the tokenized representations of the RPM images and the specific components in the representation space, as illustrated in Figure \ref{The architecture of the Representation Space.}.
This figure also underscores the necessity of tokenizing the representations of RPM images, which can significantly reduce the number of components required in the representation space. Otherwise, the representation space would have to accommodate the representations of all RPM images used for training in an undifferentiated manner, and as a result, the number of components within the space would be exceedingly large. In this paper, 256 components are set for the representation space.

The aforementioned alignment is achieved by introducing three additional loss term into Johnny's training process.
\begin{enumerate}
    \item A loss term $\ell_1$, which ensures the alignment of token $z_{ij}$ in RPM image representations with components $\{T_{\tilde{k}}\}$ within the representation space, can be expressed as:
    \begin{align}
 &\ell_1 =  \sum_{i=1}^{16}\sum_{j=1}^{N}\| z_{ij} - \text{sg}[{T}_{\tilde{k}}]\|_2^2\nonumber\\
  &\text{where} \quad \tilde{k} = \underset{k\in [1,K]}{\text{argmin}} \| {z}_{ij} - {T}_k \|_2^2 
    \end{align}
    Here, \( \text{sg}[\cdot] \) denotes the operation of stopping the gradient of a variable.
    \item A loss term $\ell_2$, which is utilized to optimize and adjust the components within the representation space, can be expressed as:
    \begin{align}
 &\ell_2 =  \sum_{i=1}^{16}\sum_{j=1}^{N}\| \text{sg}[z_{ij}] - {T}_{\tilde{k}}\|_2^2\nonumber\\
 &\text{where} \quad \tilde{k} = \underset{k\in [1,K]}{\text{argmin}} \| {z}_{ij} - {T}_k \|_2^2
    \end{align}

    \item  A mutual information loss term $\ell_3$, which facilitates the optimization of component $T_k$ within Johnny's representation space, can be expressed as:
    \begin{align}\label{111} 
 &\ell_3 =  \sum_{i=1}^{16} \| x_{i} -  \tilde{x}_i\|_2^2\nonumber\\
 &\text{where}\quad \tilde{x}_i=\text{D}({\bigcup_{j=1}^N}\{T_{\tilde k}|\tilde{k} = \underset{k\in [1,K]}{\text{argmin}} \| {z}_{ij} - {T}_k \|_2^2\})
    \end{align}
    As indicated in Equation (\ref{111}), this paper employs a mutual information supervision strategy anchored to the decoding process $D(\cdot)$. By establishing an additional decoder $D(\cdot)$ symmetric to the Representation Extraction Module, we implement a reconstruction process from the space components $\{T_{\tilde k}\}$ back to the RPM images $x_i$, thereby explicitly supervising the mutual information between the partial components $\{T_{ \tilde k}\}$ and the tokenized representations $\{z_{ij}|i\in[1,16],j\in[1,N]\}$.
\end{enumerate}

If we denote the previously defined Cross-Entropy loss term as $\ell_0$, then the complete loss function $\ell$ is designed as:
\begin{align}
 \ell=\ell_0 + \ell_1 + \lambda\cdot \ell_2 + \ell_3
    \end{align}
where $\lambda$ is set to 0.25. During the optimization of Johnny using loss function $\ell$, we apply the classical sliding window approach to manipulate components in the representation space.

\subsection{Sub-Enumeration Technique for Incorrect Options Based on the Representation Space.}

It is evident that Johnny, trained under the current loss function $\ell$, does not exhibit the claimed representation space-based sub-enumeration capability. In this paper, the loss term $\ell$ is introduced first as a foundation, since the sub-enumeration technique is strategically positioned to refine Johnny's performance after the $\ell$-based training.
Within Johnny, the sub-enumeration technique necessitates a dedicated loss function for implementation. This implies that we need to incorporate a new loss term $\ell_4$ during the later stages of Johnny's $\ell$-based training. The following is to outline the details of this new loss term.

The role of the Reasoning Module is to map the token set 
$\{z_{1j}, z_{2j}, z_{3j}, z_{4j}, z_{5j}, z_{6j}, z_{7j}, z_{8j}, z_{\alpha j}\}$ 
to the corresponding score of the $j$-th token $z_{\alpha j}$ 
in the representation of $x_{\alpha}$. where $\alpha \in [9,16]$ represents the index of the correct option. 
Figure \ref{The process by which Johnny's Reasoning Module processes RPM representations} has previously illustrated this process, which can be expressed by the equation:
\begin{align}
 \text{score}_{\alpha j} = \text{RM}(\{z_{1j}, z_{2j}, z_{3j}, z_{4j}, z_{5j}, z_{6j}, z_{7j}, z_{8j}, z_{\alpha j}\})
    \end{align}
Here, RM denotes the Reasoning Module. To introduce the concept of sub-enumeration technique to Johnny, the Reasoning Module has the additional task of scoring each component in the representation space. This can be formally expressed as:
\begin{align}\label{1111}
 \text{score}_{kj} = \text{RM}(\{z_{1j}, z_{2j}, z_{3j}, z_{4j}, z_{5j}, z_{6j}, z_{7j}, z_{8j}, T_k\})
    \end{align}
By introducing a novel loss term $\ell_4$ to impose constraints on scores $\{\text{score}_{kj}|k\in[1,K],j\in[1,N]\}$, the sub-enumeration technique can be effectively integrated into Johnny's reasoning framework. This new loss term can be interpreted as:
\begin{align}
    &{\ell _4}= - \sum_{j = 1}^{N} \log \frac{{{e^{\text{score}_{\tilde{k}j}/\tau}}}}{\sum_{k = 1}^{K} {{e^{\text{score}_{kj}/\tau}}} }\nonumber\\
    &\text{where}\quad \tilde{k} = \underset{k\in [1,K]}{\text{argmin}} \| {z}_{\alpha j} - {T}_k \|_2^2
\end{align}
Here, $\tau$ is set to 0.01 to smooth the constraint imposed by cross-entropy on the scores $\{\text{score}_{kj}|k\in[1,K],j\in[1,N]\}$.

With the explicit formulation of the loss term $\ell _4$, our objective becomes clear: Johnny first employ the Reasoning Module to score all components $\{T_{k}|k\in[1,K]\}$ in the representation space, yielding the score set $\{\text{score}_{kj}|k\in[1,K]\}$. These scores undergo temperature scaling with coefficient $\tau$ before being normalized into a probability distribution via softmax. The subsequent cross-entropy loss then enforces constraints on this probability distribution, with the optimization goal that the component $T_{\tilde{k}}$ most similar to the correct reference token $z_{\alpha j}$ (associated with the ground-truth option $x_\alpha$) should receive the highest probability mass.

Furthermore, the necessity of tokenizing the representations of RPM images and requiring the Reasoning Module to process each token independently becomes evident. This design paradigm enables the sub-enumeration loss term $\ell _4$ to be expressed as a summation over individual token-level contributions, thereby maintaining computationally efficient implementation with minimal overhead.

\section{ACT4: A Novel Extractor for Tokenized Representations of Abstract Concepts}

Evidently, the contribution of Johnny diverges from conventional RPM-solving models, which have predominantly focused on designing more sophisticated network architectures to better extract and generalize abstract concepts within RPM problems.
However, this paper also attempts to make contributions by designing a network architecture that is better suited for extracting abstract concepts. Therefore, this paper designs a novel tokenized representation extractor named Spin-Transformer, along with a lightweight version called Straw Spin-Transformer.


%

\subsection{Spin-Transformer}

Prior studies\cite{SAVIR-T,RS,Triple-CFN,capsnet} addressing RPM problems have demonstrated significant limitations in learning abstract concepts related to positional relationships between entities. These limitations are particularly pronounced in challenging benchmarks such as the OIG and D-9 problem sets from the RAVEN dataset\cite{RAVENdataset}, as well as the Neutral problems in the PGM dataset\cite{PGMdataset}.
Convolutional Neural Networks (CNNs) are indeed not particularly adept at handling positional relationships between image entities\cite{capsnet}, which is why we do not expect CNN-based abstract concept extractors (e.g., SAVIR-T, RS) to perform effectively. Given that the Transformer-Encoder also exhibits remarkable potential in image processing\cite{ViT}, we investigate its intricate mechanisms to pursue architectural breakthroughs in positional reasoning.

\subsubsection{Limitations of Transformer-Encoder in Processing RPM Images} This paper observes that during the processing of logical inputs and generation of attention-weighted outputs by the Transformer-Encoder, all output tokens are associated to varying degrees with global information originating from the logical inputs due to the self-attention mechanism \cite{Transformer,SAVIR-T,RS,ViT, Attention in natural language processing}.
In a nutshell, when denoting the logical inputs to the Transformer-Encoder as $x$, and the output tokens as $\{z_j|j\in[1,N]\}$, its inherent multi-head self-attention mechanism ensures that each component $z_j$ within the output sequence $\{z_j|j\in[1,N]\}$ implicitly encodes global contextual information derived from the entire input sequence $x$.
Precisely due to this characteristic of the self-attention mechanism, the design where the Reasoning Module in Johnny independently processes each token extracted by the Representation Extraction Module can be considered a rational architecture, enabling Johnny to achieve superior performance.

Given that each $z_j$ encodes global information about x through multi-head self-attention, we can further deduce that individual attention heads within the mechanism specialize in capturing local features of $x$. This hierarchical decomposition arises naturally, as the information capacity of any single attention head is necessarily limited to a specialized subset within the global context represented by the $z_j$.
This reveals a critical limitation of the Transformer-Encoder when processing RPM images: the parallel and independent processing across different heads in the multi-head attention mechanism implies that the model lacks Inter-head communication when handling distinct local features within RPM images. In the Transformer-Encoder, Inter-head communication between attention heads occurs solely through its feedforward block via simple linear layers. While this may suffice for natural language processing \cite{Transformer, Attention in natural language processing} and conventional vision tasks \cite{ViT, A survey of visual transformers}, it proves insufficient when addressing abstract visual reasoning challenges that necessitate capturing complex positional relationships between local features.

To enhance the performance of Transformer encoders in processing RPM images, this paper proposes a targeted enhancement strategy, which focuses on strengthening the model's ability to capture positional relationships between local features. Specifically, this is achieved by designing enhanced inter-head communication mechanisms related to positional reasoning for the Transformer-Encoder. Consequently, this paper introduces the Spin-Transformer.

\subsubsection{The architecture of Spin Block} 
\begin{figure}[htp]\centering
	\includegraphics[trim=1cm 0cm 7cm 0cm, clip, width=8.5cm]{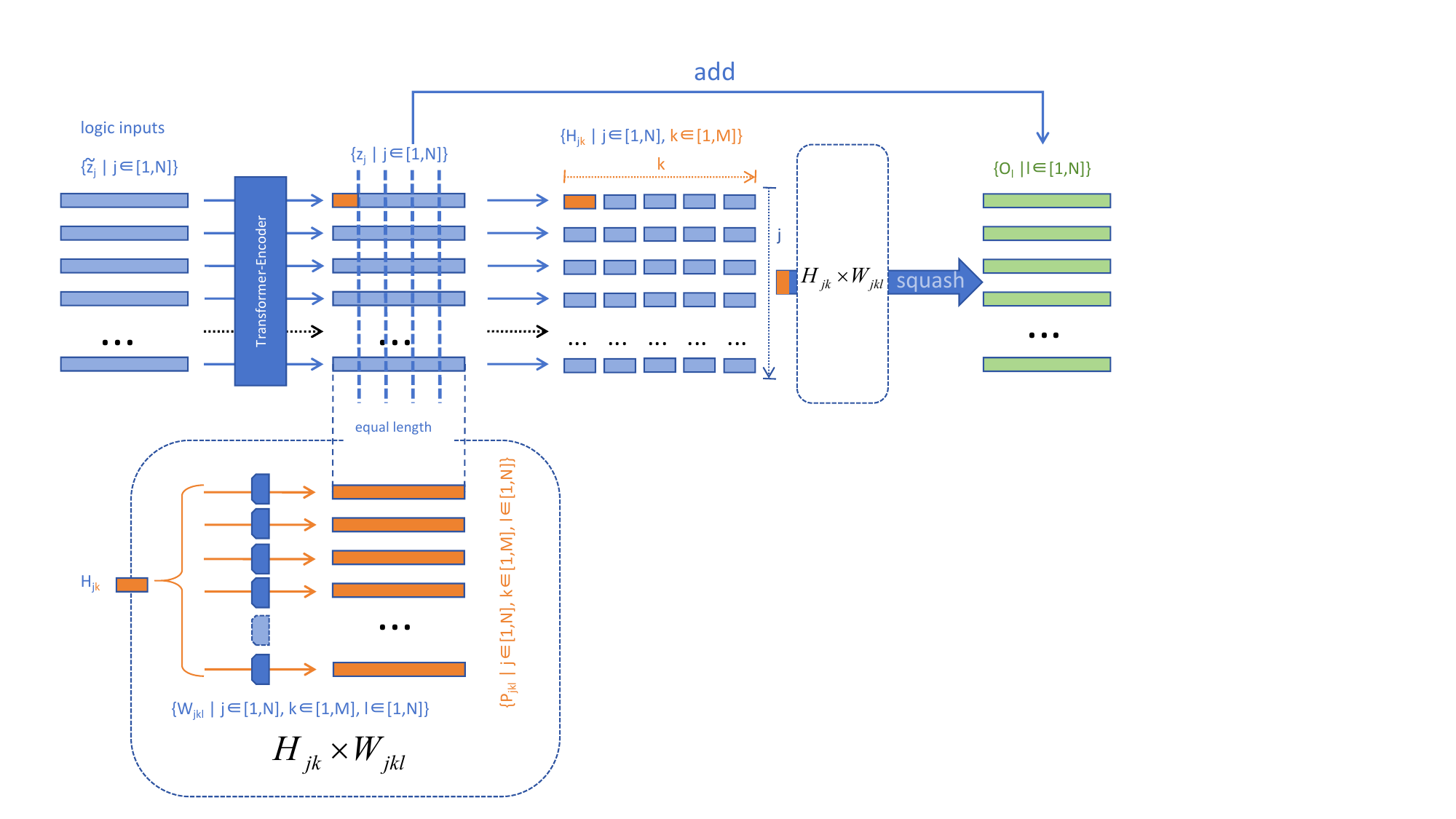}
	\caption{The feedforward process of a Spin block}
\label{pose matrix embedding}
\end{figure}

We have designed a novel block tailored for the RPM images, as illustrated in Figure \ref{pose matrix embedding}. 
The feedforward process of this block shown in the figure proceeds as follows:
\begin{enumerate}

    \item When logical inputs $x$ are split into patches $\{\tilde z_j | j\in [1,N],\tilde z_j \in R^d\}$ and pass through a single Transformer-Encoder layer, this paper segments the $N$ output tokens $\{z_j | j\in [1,N], z_j \in R^d\}$ into $M$ local vectors $\{H_{jk} | j\in [1,N], k\in [1,M], H_{jk}\in R^{d/M}\}$ of the dimension corresponding to the attention heads in the Transformer-Encoder's multi-head attention mechanism.

    \item For each local vector $H_{jk}$, we construct an associated set of mapping matrices $\{W_{jkl} | l \in [1,N]\}$. This implies the establishment of a comprehensive matrix set $\{W_{jkl} | j \in [1,N], k \in [1,M], l \in [1,N], W_{jkl} \in {R}^{(d/M) \times d}\}$, where the outer indices $(j,k)$ specifically identify their corresponding local vector $H_{jk}$, while the inner index $l \in [1,N]$ traverses all input token positions.

    \item After summation and squashing, these pose vectors are element-wise added to the vector set $\{z_j | j\in [1,N]\}$ to obtain the logical outputs $\{O_{l} | l \in [1,N], O_{l} \in R^d\}$ of this novel block. The squashing function is formulated as shown in the equation below: 
\begin{align}
    \text{squash}(\mathbf{v}) = \frac{||{\mathbf{v}}||^2}{1 + ||{\mathbf{v}}||^2} \cdot \frac{\mathbf{v}}{||{\mathbf{v}}||}
\end{align}
where the operator $||\cdot||$ is $L_2$ norm.
\end{enumerate}
Therefore, the entire calculation process of this novel block can be formulated as fallows: 
\begin{align}
    &\{ z_j | j\in [1,N]\} &=&\text{T-Encoder}(\{\tilde z_j | j\in [1,N]\})\label{q1}\\
    &\{H_{jk} | k\in [1,M]\}&=&\text{segment}( z_j )\\
    &P_{jkl}&=& H_{jk} \times W_{jkl}\\
    &O_{l} &=&\,(z_{j}|j=l) + \text{squash}(\sum_{j=1} ^{N}\sum_{k=1} ^{M}{P_{jkl}})
\end{align}
The encoding process of the single Transformer-Encoder layer is denoted as {T-Encoder} in Equation (\ref{q1}). This paper called this novel block the Spin Block.

\subsubsection{The architecture of Spin-Transformer} 
\begin{figure}[htbp]\centering
	\includegraphics[trim=6cm 0cm 12cm 0cm, clip, width=8.5cm]{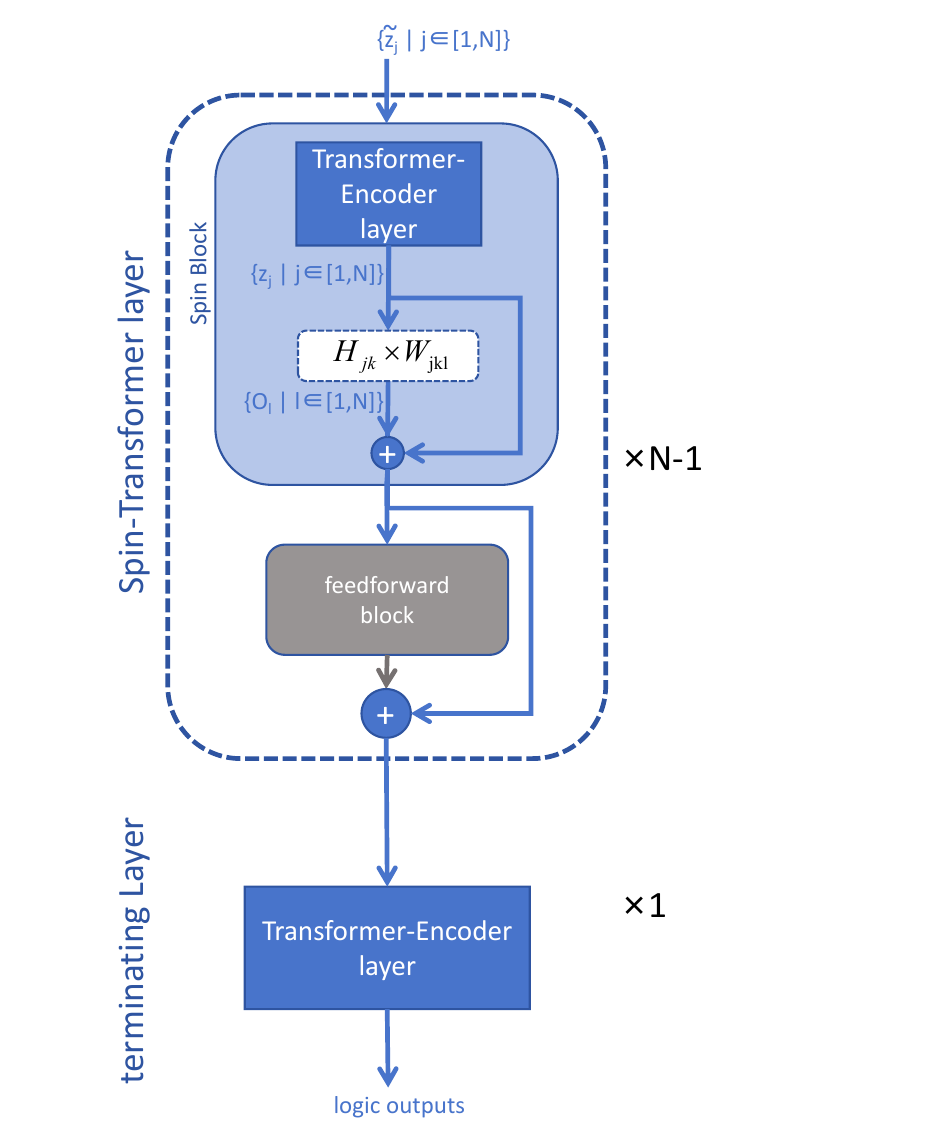}
	\caption{The architecture of the Spin-Transformer}
\label{Pose-tansformer}
\end{figure}

This Spin Block is followed by a feedforward block, forming the basic layer of the Spin-Transformer. The design of the feedforward block is consistent with that in the Transformer-Encoder architecture. Notably, both the proposed Spin Block and the feedforward block are dimension-preserving, allowing the stacking of multiple blocks to adapt to different scales of abstract reasoning problems. The Spin-Transformer with $N$ logical layers is depicted in the Figure \ref{Pose-tansformer}. 
\begin{enumerate}
    \item When $N=1$, the Spin-Transformer collapses into a regular Transformer-Encoder. In Transformer-Encoder attention weights in the Transformer-Encoder are distributed over lower-level representations and computed top-down.

    \item  In the Spin-Transformer with $N>1$, similarity weights for local representations can be calculated at multiple levels and scales from the outside in. This is ensured by the fact that each logical layer in the Spin-Transformer is succeeded by a standard Transformer-Encoder layer, which can be present in the subsequent logical layer or the terminating layer, responsible for recalculating the similarity weights.
\end{enumerate}
Spin-Transformer introduces the computation of pose vectors $P_{jkl}$, thereby enhancing the induction of pose relationships between local representations. However, it also introduces significant computational overhead due to the requirement of optimizing parameters for the matrix set $\{W_{jkl}\}$

\subsection{Straw Spin-Transformer}

During deployment, Spin-Transformer imposes substantial computational demands. Notably, the process of establishing and optimizing the mapping matrices $\{W_{jkl}\}$ significantly exacerbates resource consumption. Addressing this challenge, we investigate lightweight design strategies for Spin-Transformer. The core mechanism revolves around computing pose vectors $P_{jkl}$, which are then utilized to modify and refine logical inputs $z_{j}$. This process can be expressed as:
\begin{align}\label{oll}
O_l = z_j + squash\left(\sum_{j=1}^{N}\sum_{k=1}^{M} {H_{jk\_}} \times W_{jkl} \right)
\end{align}
It is worth noting that the placeholders marked with ``\_" in the formula represent numbering for on-demand broadcasting.

It is evident from Equation (\ref{oll}) that multi-level summation and mapping operations incur significant computational overhead. By designing a parameter sharing mechanism, we aim to eliminate redundant logical summation and mapping operations, thereby achieving the goal of reducing computational costs. From the logical hierarchy perspective in Equation (\ref{oll}), only the indexing at the $j$-th level allows for adjustment, while other logical levels have stringent mapping requirements. In other words, this paper posits that the matrix mappings between equivalent local vectors across different tokens warrant weight sharing. The targeted expression for the adjusted matrix mapping is as follows:
\begin{align}\label{ol}
O_l = z_j + squash\left(\sum_{k=1}^{M} {H_{k\_}} \times W_{kl} \right)
\end{align}

\subsubsection{Lightweight Methodology of Spin Block} This paper aims to simplify the Spin-Transformer based on the Equation (\ref{o3}-\ref{o6}). 
\begin{align}
O_l =& z_j + squash\left(\sum_{j=1}^{N}\sum_{k=1}^{M} {H_{jk\_}} \times W_{jkl} \right)\label{o3}\\
\tilde O_l =& z_j + squash\left(\sum_{j=1}^{N}\sum_{k=1}^{M} {H_{jk\_}} \times (\omega_j \cdot \tilde W_{\_kl}) \right) \label{o4}\\
=& z_j + squash\left(\sum_{k=1}^{M} (\sum_{j=1}^{N} \omega_j \cdot {H_{jk\_}}) \times \tilde W_{kl} \right)\label{o5}\\
\tilde=& z_j + squash\left(\sum_{k=1}^{M} \text{Attention}({H_{jk}}) \times \tilde W_{kl} \right)\label{o6}
\end{align}

\begin{enumerate}

    \item In Equation (\ref{o4}), we reparameterize each mapping matrix $W_{jkl}$ as a scaled product of a shared base matrix $\tilde W_{kl}\in R^{(d/m)\times d}$ and $N$ learnable coefficients $\{\omega_j|j\in[1,N]\}$ (i.e., $W_{jkl} \tilde= \omega_j \cdot \tilde W_{\_ kl}$). This coefficient-based reparameterization significantly reduces the number of trainable matrices from $O(N)$ to $O(1)$.

    \item In Equation (\ref{o5}), we posit that adjusting the binding order among $\omega_j$, $H_{jk}$, and $\tilde W_{ kl}$ - specifically by first combining $\omega_j$ with $H_{jk}$ - and interpreting the summation $\sum_{j=1}^{N}\omega_j \cdot H_{jk\_}$ as a learnable weighted summation process for $H_{jk}$, constitutes a more technically sound approach than treating $\omega_j$ as independent optimizable weights paired with $\tilde W_{kl}$.

    \item In Equation (\ref{o6}), we posit that, instead of treating $\sum_{j=1}^{N}\omega_j \cdot H_{jk}$ as a learnable weighted summation process for $H_{jk}$, substituting it with an attention-weighted summation mechanism for $H_{jk}$ demonstrates greater technical elegance and computational efficiency.

    \item This paper proposes a feasible solution that replaces the weighted summation process with an attention mechanism:  (1) first introducing a learnable embedding vector $z_0$, where $z_{0} \in R^{d}$; (2) then defining the attention heads corresponding to $z_0$ as $\{H_{0k}|k \in [1,M], H_{0j} \in R^{(d/M)} \}$; and (3) finally computing the cross-attention results with $H_{0k}$ as the query and $\{H_{jk}|j\in [1,N]\}$ as the key-value pairs, thereby substituting the original weighted summation operation $\sum_{j=1}^{N}\omega_j \cdot H_{jk}$.

    \item The multi-head cross-attention results between $H_{0k}$ and $\{H_{jk}|j\in[1,N]\}$ can be interpreted as a weighted summation performed over the set $\{H_{jk} | j\in [1,N]\}$, where the weights are determined by similarity scores computed between each $H_{jk}$ in the set and $H_{0k}$. Therefore, the cross-attention mechanism can serve as an alternative to the learnable weighted summation process $\sum_{j=1}^{N}\omega_j \cdot H_{jk}$.
\end{enumerate}
This process compresses multiple mapping matrices into a cross-attention computation framework. This enables the feedforward process of the Spin Block to be updated as:
\begin{align}
    &\{ z_j | j\in [1,N]\} &=&\text{T-Encoder}(\{\tilde z_j | j\in [1,N]\})\label{x1}\\
    &\text{Introduce}&:&z_0\in R^d\\
    &\{H_{jk} | k\in [1,M]\}&=&\text{segment}(z_j)\\
    &H_{0k} &=& \text{cross-attention} \left\{ 
  \substack{
    \text{query: } H_{0k} \\ 
    \text{key-value: } \{ H_{jk} | j \in [1,N] \}
  } 
\right\}\\
    &P_{kl}&=& H_{0k} \times \tilde W_{kl}\\
    &O_l &=&\,(z_{j}|j=l) + \text{squash}(\sum_{k=1} ^{M}{P_{kl}})\label{x2}
\end{align}
However, the Spin Block with the aforementioned feedforward structure still exhibits significant limitations.

\subsubsection{The architecture of Straw Spin Block}
It is critical to note that, as evident in Equations (\ref{x1})-(\ref{x2}), this methodology necessitates the incorporation of additional cross-attention mechanisms between the standard Transformer-Encoder layer and matrix mapping operations. However, this incorporation extends the training duration, directly conflicting with the lightweight design objective. To resolve this contradiction, we propose a novel Transformer-Encoder architecture incorporating an auxiliary masking mechanism. This architecture integrates multi-head cross-attention computation into the multi-head self-attention mechanism within the Transformer-Encoder.

\begin{figure}[htp]\centering
	\includegraphics[width=8.5cm]{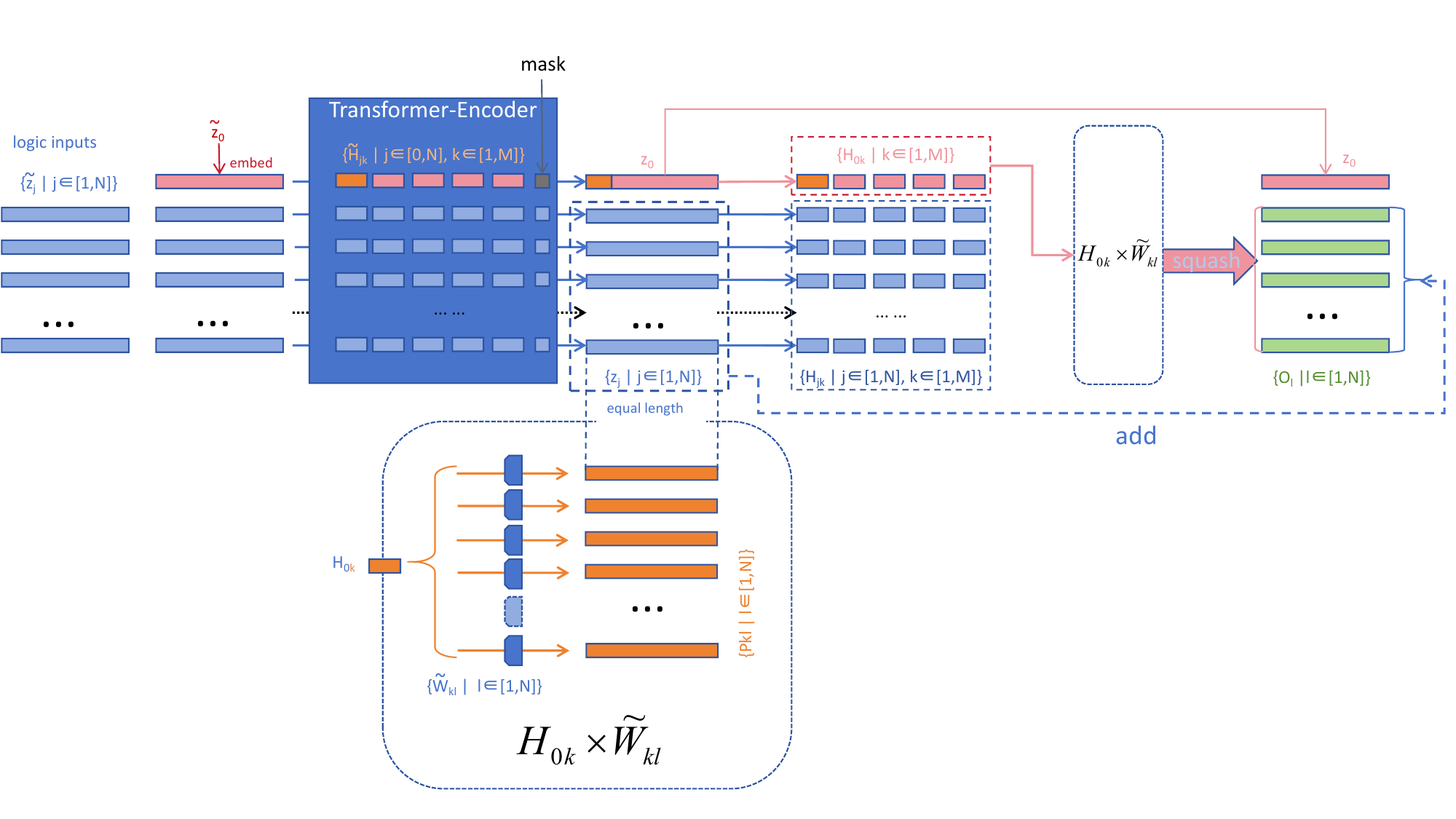}
	\caption{The feedforward process of a Straw Spin Block}
\label{Straw-Pose-Transformer layer}
\end{figure}

Specifically, this paper aims to implement the novel Spin Block architecture as illustrated in Figure \ref{Straw-Pose-Transformer layer}.
This architecture is called Straw Spin Block.
The feedforward steps in the figure can be interpreted as:
\begin{enumerate}

    \item This paper introduces a learnable vector, denoted as
    $\tilde z_0 \in R^d$, to embed the set of logical inputs $\{\tilde z_{j}|j \in [1,N], z_{j} \in R^d \}$, thereby deriving the updated logical input set $\{\tilde z_{j}|j \in [0,N], z_{j} \in R^d \}$.

    \item The logical inputs $\{\tilde{z}_j | j \in [0,N]\}$ is initially decomposed into multiple attention heads $\{\tilde{H}_{jk}| j \in [0,N], k \in [1,M], \tilde{H}_{jk} \in {R}^{(d/M)}\}$ by a single Transformer-Encoder layer. These attention heads undergo parallel computation through the multi-head attention mechanism and feedforward block within the Transformer-Encoder layer, generating the logical outputs $\{z_j | j \in [0,N], z_j \in {R}^d\}$. Thereafter, these logical outputs are repartitioned into localized vector sets $\{H_{jk}| j \in [0,N], k \in [1,M], H_{jk} \in {R}^{(d/M)}\}$ according to the original attention head dimensions. During this processing pipeline, a mask is applied to the self-attention weights at the position corresponding to $\tilde z_0$. 
    
    \item This masking operation prevents the value mapping of $\tilde H_{0k}$ from affecting the attention computation results. The masking mechanism ensures that $ H_{0k}$ is generated exclusively through the weighted summation of $\{\tilde H_{jk}|j \in [1,N]\}$, while explicitly prohibiting $\tilde H_{0k}$ from participating in this calculation process. This masked attention can be expressed as Figure \ref{MAsked attention}.
    \begin{figure}[htp]\centering
	\includegraphics[trim=2.5cm 0cm 6.4cm 0cm, clip, width=8.5cm]{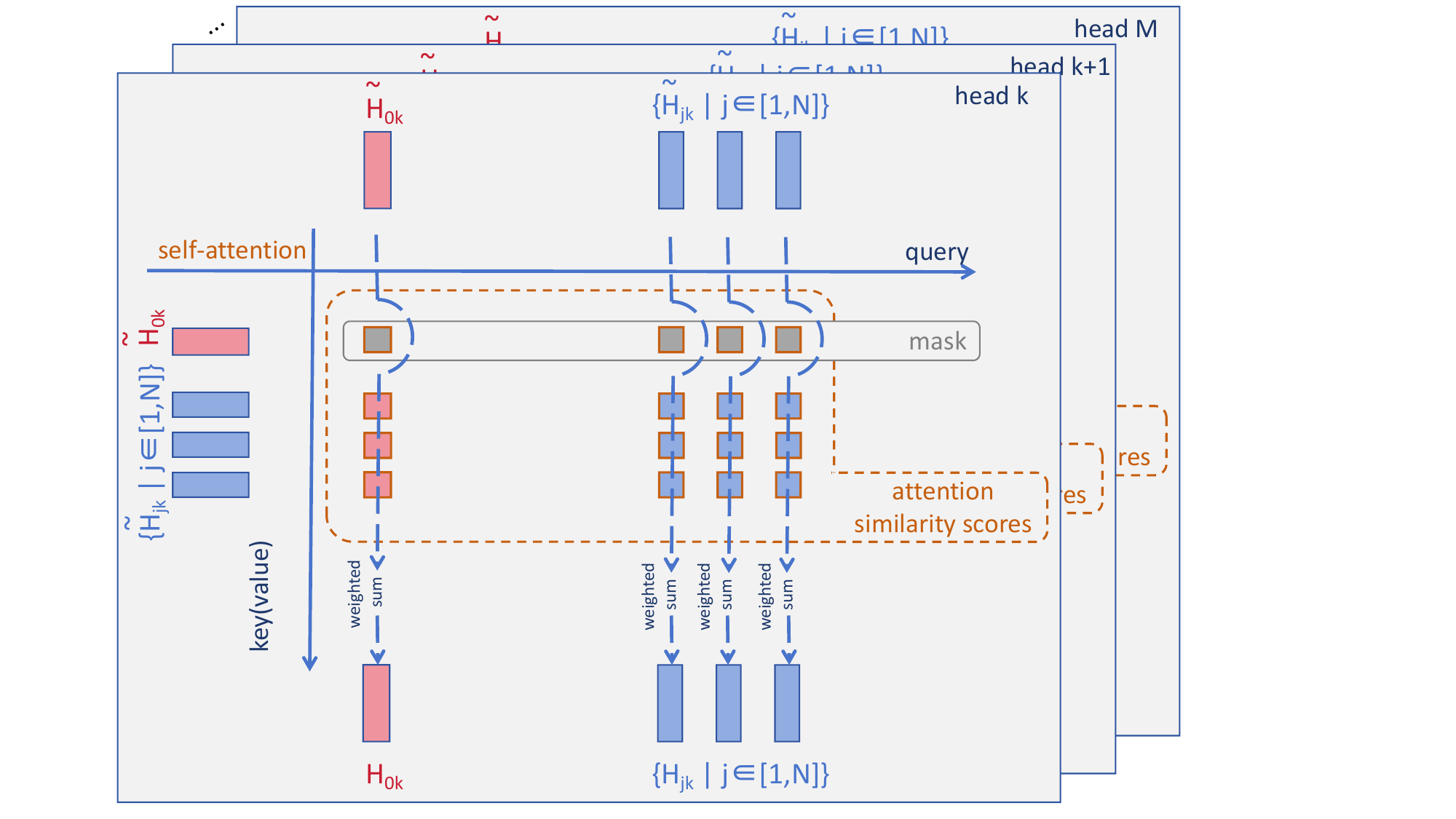}
	\caption{Masked attention}
    \label{MAsked attention}
    \end{figure}

    \item  Due to the combined effects of the multi-head self-attention and masking operations, the vector $\tilde z_0$ aggregates contextual information from the logical inputs $\{\tilde z_{j}|j \in [1,N] \}$, and is ultimately transformed into the global vector $z_0$. Within this global vector, we establish a set of mapping matrices $\{\tilde W_{kl} |k\in[1,M], l\in[1,N], \tilde W_{kl}\in R^{(d/m)\times d}\}$ for each attention head $\{H_{0k}|k \in [1,M] \}$ contained therein.

    \item  The pose vectors $\{P_{kl} | k\in[1,M], l\in[1,N], P_{kl}\in {R}^{d}\}$ are achieved by multiplying each attention head $H_{0k}$ with the mapping matrices $\{\tilde{W}_{kl} | l\in [1,N]\}$:
    \begin{align}
        P_{kl} = H_{0k} \times \tilde{W}_{kl}
    \end{align}
    After summation and squashing, these pose vectors are element-wise added to the vector set $\{z_j | j\in [1,N]\}$ to obtain the outputs of the Straw Spin Block $\{O_l | \,l\in [1,N], O_l\in R^d\}$. 

    \item The outputs $\{O_l|l\in[1,N]\}$ are concatenated with the global vector $z_0$. Thereafter, the augmented set $\{z_0, O_l|l\in[1,N]\}$ is fed into the subsequent block for further processing. 
\end{enumerate}
The entire calculation process of Straw Spin Block can be expressed as fallows: 
\begin{align}
    &\text{Introduce}&:&\tilde z_0\in R^d\\
    & \{\tilde z_j | j\in [0,N]\}&=&\text{concatenate}({\tilde z_0,\{\tilde z_j | j\in [1,N]\}})\\
    &\{ z_j | j\in [0,N]\} &=&\text{MT-Encoder}(\{\tilde z_j | j\in [0,N]\})\label{p1}\\
    &\{H_{0k} | k\in [1,M]\}&=&\text{segment}( z_0 )\\
    &P_{kl}&=& H_{0k} \times \tilde W_{kl}\\
    &O_l &=&\,(z_{j}|j=l) + \text{squash}(\sum_{k=1} ^{M}{P_{kl}})
\end{align}
In the Equation (\ref{p1}), the encoding process of the Masked Transformer-Encoder to is denoted as {MT-Encoder}.

\subsubsection{The architecture of Straw Spin-Transformer} 
\begin{figure}[htbp]\centering
	\includegraphics[trim=5cm 0cm 15cm 0cm, clip, width=7cm]{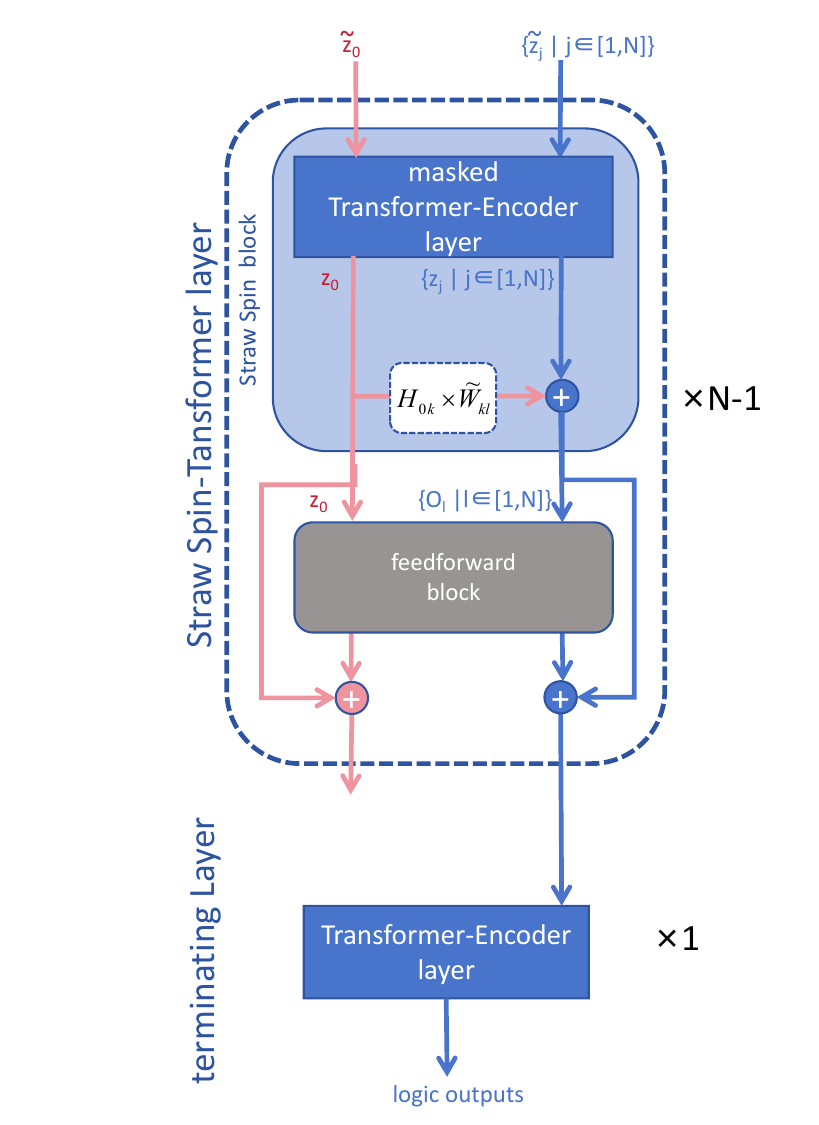}
	\caption{The architecture of the Straw Spin-Transformer}
\label{Straw-Pose-Transformer}
\end{figure}

When the Straw Spin Block with the aforementioned architecture is stacked with a feedforward block, it forms a basic layer of the Straw Spin-Transformer. As illustrated in Figure \ref{Straw-Pose-Transformer}, the Straw Spin-Transformer with $N$ logical layers is constructed from $N-1$ stacked basic layers and a terminating layer. Notably, when the Straw Spin Block is succeeded by a terminating layer, the generated logical outputs $\{O_l| l \in [1,N]\}$ bypasses the concatenation operation with the global vector $z_0$ and is directly fed into the terminating layer for subsequent processing.

\subsection{Contribution of Spin-Transformer and Straw Spin-Transformer} The Spin-Transformer and Straw Spin-Transformer architectures can replace the Transformer-Encoder structure in Johnny to further enhance its performance. Furthermore, this paper notes that several high-performing RPM-solving models \cite{RS,Triple-CFN} have incorporated the Transformer-Encoder architecture into their structural designs. Correspondingly, Spin-Transformer and Straw Spin-Transformer can further enhance their performance by substituting the Transformer-Encoder components within these models.

Notably, despite the significant reduction in computational overhead achieved by the Straw Spin-Transformer, there remains a non-negligible computational cost. Consequently, when integrating Spin-Transformer with Johnny, the system must forego the construction of representation spaces and the application of sub-enumeration techniques.

\section{Experiment}
All experiments conducted in this paper were programmed in Python, utilizing the PyTorch\cite{Pytorch} framework. Johnny operates in four distinct configurations:
 
\begin{enumerate}
    \item {Baseline Configuration}:  
    Johnny, equipped with the Representation Extraction Module and Reasoning Module, undergoes optimization and training constrained by the loss term $\ell_0$. In this configuration, Johnny shares the same operational paradigm with traditional RPM-solving models: it assigns a validity probability to each candidate in the RPM problem's option pool, thereby enabling rudimentary reasoning capabilities.

    \item {Augmented Representation Space}:
    By constructing an auxiliary representation space atop the existing Representation Extraction and Reasoning Modules, the encoding capability of the Representation Extraction Module is enhanced. This configuration adopts the composite loss function $\ell$ as its optimization constraint.

    \item {Advanced Configuration with Sub-enumeration}:
    When integrating the Representation Extraction Module, Reasoning Module, augmented representation space, and sub-enumeration technique, the system necessitates the combined loss function $\ell + \ell_4$ for effective training. In this configuration, Johnny leverages the representation space to rectify incorrect option configurations within RPM instances, thereby unlocking its potential.

    \item {Spin-Transformer Backbone Variant}:
    In this configuration, while maintaining the core Representation Extraction and Reasoning Modules, the Transformer-Encoder backbone architecture is replaced with Spin-Transformer. The primary loss function remains $\ell_0$. The paper posits that Spin-Transformer, compared to the Transformer-Encoder, incorporates a more robust inter-head communication mechanism primarily tasked with abstracting positional relationships among local features, thereby further enhancing Johnny's performance.

\end{enumerate}
Therefore, this paper conducted experiments on the RAVEN \cite{RAVENdataset}, I-RAVEN\cite{I-RAVEN} and PGM \cite{PGMdataset} datasets, centered around these four configurations. The experiment results were recorded in the table’s entries titled Johnny ACT1, Johnny ACT2, Johnny ACT3, and Johnny ACT4, respectively. The method employing Spin-Transformer is denoted as ACT4, while the Straw Spin-Transformer variant is denoted as ACT4$^-$.

\subsection{Experiment on  RAVEN}
This paper first conducted experiments on the RAVEN and I-RAVEN datasets for configurations Johnny ACT1-3. For a fair comparison, we used the same settings and equipment as RS-Tran\cite{RS} and Triple-CFN\cite{Triple-CFN}, including data volume, optimizer parameters, data augmentation, batch size.
The results of the experiment are presented in the Table \ref{RAVEN_IRAVEN_Results}
\begin{table}[h]
\caption{Reasoning accuracies on RAVEN and I-RAVEN.}
\label{RAVEN_IRAVEN_Results}
\centering
\resizebox{\linewidth}{!}{
\begin{tabular}{cccccccccc}
\toprule
\toprule
&\multicolumn{8}{c}{Test Accuracy(\%)}& \\
\cmidrule{2-9}
Model&Average&Center&2 $\times$ 2 Grid&3 $\times$ 3 Grid&L-R&U-D&O-IC&O-IG \\
\midrule
SAVIR-T \cite{SAVIR-T}&94.0/98.1&97.8/99.5&94.7/98.1&83.8/93.8&97.8/99.6&98.2/99.1&97.6/99.5&88.0/97.2\\
\cmidrule{2-9}
SCL \cite{SCL, SAVIR-T}&91.6/95.0&98.1/99.0&91.0/96.2&82.5/89.5&96.8/97.9&96.5/97.1&96.0/97.6&80.1/87.7\\
\cmidrule{2-9}
MRNet \cite{MRNet}&96.6/-&-/-&-/-&-/-&-/-&-/-&-/-&-/-\\
\cmidrule{2-9}
RS-TRAN\cite{RS}&{98.4}/98.7&99.8/{100.0}&{99.7}/{99.3}&{95.4}/96.7&99.2/{100.0}&{99.4}/99.7&{99.9}/99.9&{95.4}/95.4 \\
\cmidrule{2-9}
Triple-CFN\cite{Triple-CFN}&98.9/{99.1}&100.0/{100.0}&99.7/{99.8}&96.2/{97.5}&99.8/{99.9}&99.8/{99.9}&99.9/99.9&97.0/{97.3} \\
\midrule
Johnny ACT1 & 98.6/99.0 & 100.0/100.0 & 99.4/99.5 & 95.7/96.9 & 99.7/99.8 & 99.7/99.8 & 99.8/99.8 & 96.0/97.3 \\
\cmidrule{2-9}
Johnny ACT2 & 98.8/99.2 & 100.0/100.0 & 99.6/99.7 & 96.1/97.4 & 99.8/99.9 & 99.8/99.9 & 99.8/99.8 & 96.7/97.9 \\
\cmidrule{2-9}
Johnny ACT3 & 99.4/99.6 & 100.0/100.0 & 99.6/99.7 & 98.1/99.1 & 99.8/100.0 & 99.8/100.0 & 99.9/99.9 & 98.5/99.3 \\
\cmidrule{2-9}
Johnny ACT4&99.4/{99.5}&100.0/{100.0}&99.8/{99.9}&97.9/{98.2}&100.0/{100.0}&100.0/{100.0}&100.0/100.0&98.0/{98.6} \\
\cmidrule{2-9}
Johnny ACT4$^-$&99.4/99.5&100.0/{100.0}&99.6/{99.9}&97.7/98.1&100.0/{100.00}&100.0/{100.0}&100.0/100.0&97.8/98.3 \\
\bottomrule
\bottomrule
\end{tabular}
}
\end{table}
As demonstrated in Table \ref{RAVEN_IRAVEN_Results}, the Johnny ACT1-4 have achieved impressive results on both the RAVEN and I-RAVEN datasets, thereby substantiating the effectiveness of the methodologies embedded in the design philosophy of the Johnny ACT1-3.

To demonstrate the superiority of the inter-head communication mechanism modification in Spin-Transformer for capturing positional relationships between local features when addressing RPM problems, we replaced the original Transformer-Encoder modules in state-of-the-art RPM-solving models (including RS-TRAN and Triple-CFN) with our proposed (Straw) Spin-Transformer. The experiment results are recorded in Table \ref{RAVEN_IRAVEN_Results_}. We considered that a single (Straw) Spin-Transformer Layer incorporates a two-stage representation processing mechanism (e.i. the attention mechanism and position-matrix mapping mechanism), which involves more computational components compared to the standard Transformer-Encoder. Therefore, when conducting the module replacement experiments with RS-Tran \cite{RS} and Triple-CFN \cite{Triple-CFN}, we configured the Spin-Transformer layers to half the number of the original Transformer-Encoder layers. This adjustment ensures computational parity while enabling more effective validation of the enhanced reasoning capabilities introduced by Spin-Transformer's architectural innovations.
\begin{table}[htbp]
\caption{Reasoning Accuracies of (Straw) Spin-Transformer on RAVEN and I-RAVEN.}
\label{RAVEN_IRAVEN_Results_}
\centering
\resizebox{\linewidth}{!}{
\begin{tabular}{cccccccccc}
\toprule
\toprule
&\multicolumn{8}{c}{Test Accuracy(\%)}& \\
\cmidrule{2-9}
Model&Average&Center&2 $\times$ 2 Grid&3 $\times$ 3 Grid&L-R&U-D&O-IC&O-IG \\
\midrule
RS-TRAN\cite{RS}&{98.4}/98.7&99.8/{100.0}&{99.7}/{99.3}&{95.4}/96.7&99.2/{100.0}&{99.4}/99.7&{99.9}/99.9&{95.4}/95.4 \\
\cmidrule{2-9}
RS-Tran+ACT4&99.1/{99.5}&100.0/{100.0}&99.7/{99.9}&97.5/{98.2}&99.9/{100.0}&99.9/{100.0}&99.8/99.9&97.1/{98.6} \\
\cmidrule{2-9}
RS-Tran+ACT4$^-$&99.0/99.4&100.0/{100.0}&99.7/{99.8}&97.1/98.1&99.9/{100.0}&99.9/{100.0}&99.7/{99.9}&96.9/98.2 \\
\midrule
Triple-CFN\cite{Triple-CFN}&98.9/{99.1}&100.0/{100.0}&99.7/{99.8}&96.2/{97.5}&99.8/{99.9}&99.8/{99.9}&99.9/99.9&97.0/{97.3} \\
\cmidrule{2-9}
Triple-CFN+ACT4&99.4/99.5&100.0/{100.0}&99.8/{99.9}&97.9/{98.1}&99.9/{100.0}&99.9/{100.0}&99.9/{99.9}&97.8/{98.6} \\
\cmidrule{2-9}
Triple-CFN+ACT4$^-$&99.3/99.5&100.0/{100.0}&99.8/{99.8}&97.6/{98.1}&99.9/{100.0}&99.9/{100.0}&99.9/{99.9}&97.5/{98.3} \\
\bottomrule
\bottomrule
\end{tabular}
}
\end{table}

Table \ref{RAVEN_IRAVEN_Results_} demonstrates that (Straw) Spin-Transformer has delivered impressive performance improvements to RS-Tran and Triple-CFN on RAVEN sub-problems such as 3$\times$3 Grid and OI-G, which require intensive analysis of positional relationships between entities. This validates the effectiveness of the novel mechanisms proposed in (Straw) Spin-Transformer. As a lightweight version of Spin-Transformer, Straw Spin-Transformer does not significantly degrade performance, validating the effectiveness of our lightweight approach.

\subsection{Experiment on  PGM}
In this section, experiments were conducted using the identical configuration on the PGM dataset, with the results documented in Table \ref{PGM_Results}.
\begin{table}[htbp]
\caption{Reasoning accuracies of models on PGM.}
\label{PGM_Results}
\centering
\begin{tabular}{ccc}
\toprule
Model&Test Accuracy(\%) \\
\midrule
{CoPINet\cite{CoPINet}}&56.4\\
\midrule
SAVIR-T \cite{SAVIR-T}&91.2\\
\midrule
SCL \cite{SCL, SAVIR-T}&88.9\\
\midrule
MRNet \cite{MRNet}&94.5\\
\midrule
RS-CNN\cite{RS}&82.8\\
\midrule
RS-TRAN\cite{RS}&{97.5}\\
\midrule
Triple-CFN\cite{Triple-CFN}&{97.8}\\
\midrule
\midrule
Johnny ACT1&{97.9}\\
\midrule
Johnny ACT2&{98.2}\\
\midrule
Johnny ACT3&\textbf{99.0}\\
\midrule
Johnny ACT4&{98.8}\\
\midrule
Johnny ACT4$^-$&{98.8}\\
\midrule
\midrule
RS-TRAN+ACT4&{-}\\
\midrule
RS-TRAN+ACT4$^-$&{98.6}\\
\midrule
Triple-CFN+ACT4&{98.6}\\
\midrule
Triple-CFN+ACT4$^-$&{98.7}\\
\bottomrule
\end{tabular}
\end{table}
As observed in Table \ref{PGM_Results}, we were unable to successfully replace the Transformer-Encoder in RS-Tran with Spin-Transformer for PGM problems. This failure stems from the fact that solving PGM tasks with RS-Tran requires doubling the overall dimensions of model parameter matrices compared to RAVEN problem setups. Such dimensions expansion significantly amplifies the computational burden associated with Spin-Transformer's inherent matrix mapping operations. This challenge directly motivated the development of our lightweight variant, Straw Spin-Transformer.

The paper also conducted experiments on the generalization sub-problems of PGM, with the results documented in Table \ref{Generalization_PGM_2}.
\begin{table}[h]
\caption{Generalization Results of Valen in PGM.}
\label{Generalization_PGM_2}
\centering
\resizebox{\linewidth}{!}{
\begin{tabular}{cccccccc}
\toprule
& \multicolumn{7}{c}{Dataset} \\
\cmidrule{2-8}
\makecell{Model/\\Task} & Interpolation & Extrapolation & \makecell{Held-out\\Attribute\\shape-colour} & \makecell{Held-out\\Attribute\\line-type} & \makecell{Held-out\\Triples} & \makecell{Held-out\\Pairs of\\Triples} & \makecell{Held-out\\Attribute\\Pairs} \\
\midrule
\makecell{MRNet\cite{MRNet}} & {68.1} & 19.2 & 16.9 & 30.1 & 25.9 & {55.3} & {38.4} \\
\midrule
\makecell{RS-Tran\cite{RS}} & {77.2} & 19.2 & 12.9 & 24.7 & 22.2 & {43.6} & {28.4} \\
\midrule
\makecell{Triple-CFN\cite{Triple-CFN} } & {80.4} & 18.4 & 12.6 & 25.2 & 22.0 & {44.5} & {29.2} \\
\midrule
\midrule
Johnny ACT1 & 81.0 & 18.7 & 12.5 & 26.2 & 23.1 & 45.1 & 29.3 \\
\midrule
\makecell{Johnny ACT2} & 82.5 & 18.6 & 13.5 & 26.1 & 23.0 & 45.0 & 29.8 \\
\midrule
\makecell{Johnny ACT3} & \textbf{87.3} & 18.8 & 13.3 & 27.5 & 23.8 & 45.5 & 30.5 \\
\midrule
\makecell{Johnny ACT4} & 83.2 & 18.2 & 13.4 & 26.7 & 23.6 & 45.4 & 30.2 \\
\midrule
\makecell{Johnny ACT4$^-$} & 82.8 & 18.4 & 13.1 & 26.3 & 23.2 & 44.9 & 29.7 \\
\bottomrule
\end{tabular}
}
\end{table}
It can be observed from the table that the methodology proposed in this paper, which establishes a representation space and incorporates sub-enumeration techniques, has achieved remarkable progress in addressing the Interpolation task within the PGM image attribute generalization problem. However, its performance appears somewhat limited when confronted with progressive generalization tasks. However, this paper argues that conclusively determining model superiority near 50\% accuracy is problematic, as such performance levels inherently indicate limitations. Comparative evaluations under these constraints lack validity. Therefore, the progressive generalization sub-problem of PGM under sparse supervision remains an open challenge.

\subsection{Experiment on PGM Generalization Sub-problems.}

When considering the availability of more supervisory signals, certain progressive pattern generalization sub-problems within PGM appear solvable. This paper emphasizes the metadata associated with each PGM instance, which comprehensively documents all progressive patterns embedded within that instance. By incorporating this metadata as additional supervisory signals into PGM generalization tasks, enumeration techniques and representation space construction methodologies can  make more contributions.

The approach proposed in this paper to solve PGM generalization sub-problems using representation space and enumeration techniques is straightforward: When there exists a representation space that enumerates all possible manifestations of metadata, the matrix pattern extracted by Johnny for the deduction matrix is mapped, guided by metadata cues, to corresponding components within this representation space. This methodology, analogous to the approach illustrated in Figure \ref{The architecture of the Representation Space.}, demonstrates potential for further contributions.

Specifically, we enumerate all possible manifestations of metadata and establish a representation space $\{Y_\beta \mid \beta \in [1,L]\}$ with a learnable component count matching that of the metadata manifestations. Subsequently, four matrix pattern vectors, denoted as $\{ q_{ j \alpha} | \alpha \in [1,4] \}$, are derived from the image representation $\{ z_{1j}, z_{2j}, z_{3j}, z_{4j}, z_{5j}, z_{6j}, z_{7j}, z_{8j} \}$ following the procedure illustrated in Figure \ref{The architecture of the Pattern Extractor.}. Two of these vectors will be optimized using a cross-entropy loss function, guided by {metadata}, to align with corresponding components in the latent representation space. The design of selecting two vectors from $\{q_{j \alpha } | \alpha \in [1,4]\}$ to separately align with the representation space stems from the fact that each PGM instance contains two sets of decoupled progressive patterns - In other words, each PGM instance is guided by two metadata directives, necessitating the selection of two vectors from $\{q_{j \alpha } | \alpha \in [1,4]\}$ to receive guidance and constraints from the metadata.
This aligning can be achieved by a new loss function:
\begin{align}
    &{\ell _{5}}= -\sum_{\alpha=1}^{2}  \sum_{\tilde \beta = 1}^{L} \text{metadata}_{{\alpha\tilde \beta}}\cdot\log \frac{{{e^{({\bar q_{\alpha}} \cdot {  Y_{\tilde \beta}})/\tau}}}}{\sum_{\beta = 1}^{L} {{e^{({\bar q_{\alpha} } \cdot {Y_{\beta}})/\tau}}} }\label{act3}\nonumber\\
    &\text{where}\quad \bar q_{\alpha}=\frac{1}{N}\sum_{j=1}^{N}q_{ j\alpha}
\end{align}
Here, $\tau$ represents an optimizable positive-valued temperature parameter, initialized to $1 \times 10^{-6}$.

When Johnny uses $\ell_0 + \ell_5$ as the training loss function, the reasoning accuracy Johnny achieves on the generalization sub-problems of PGM is recorded in Table \ref{Generalization_PGM_3}.
\begin{table}[h]
\caption{Generalization Results of Johnny ACT1 on PGM with the Assistance of Metadata}
\label{Generalization_PGM_3}
\centering
\begin{tabular}{ccc}
\toprule
&\multicolumn{2}{c}{Accuracy(\%)}\\
Dataset&Johnny ACT1&\\
\midrule
Interpolation &\textbf{92.2} \\
\midrule
Extrapolation &13.0\\
\midrule
Held-out Attribute shape-colour &14.0\\
\midrule
Held-out Attribute line-type&27.0 \\
\midrule
Held-out Triples&28.5\\
\midrule
Held-out Pairs of Triples &\textbf{98.0} \\
\midrule
Held-out Attribute Pairs&\textbf{98.0}  \\
\bottomrule
\end{tabular}
\end{table}

It can be observed from the Table \ref{Generalization_PGM_2} and \ref{Generalization_PGM_3} that the enumeration framework and representation space construction methodology proposed in this work make consistent contributions across multiple sub-problems of the RPM. However, it is crucial to emphasize that any operation involving enumeration inherently implies assumptions about the boundedness, discreteness, and enumerability of the objects being processed. While the Representation Extraction Module and corresponding representation space developed in Sections \ref{ACT3} achieved discretization of RPM image attributes - thereby obviating the need to assume enumerability during their construction - we must acknowledge that our approach to solving PGM generalization problems through metadata inherently relies on the enumerability of metadata. This fundamental dependency reveals the methodological limitations of employing metadata-driven approaches for RPM problem-solving.

\section{Conclusion}

This paper systematically explores innovative architectural designs based on representation learning and their applications in RPM tasks, centered around the core objective of enhancing the abstract reasoning capabilities of deep learning models. By introducing the Johnny architecture, Spin-Transformer network structure, and sub-enumeration technology, this study has achieved significant advancements in both theoretical breakthroughs and empirical analyses, providing a new paradigm for addressing the limitations of deep learning models in the field of abstract reasoning. The contributions are as follows:
\begin{enumerate}
\item {Johnny Architecture:}
By constructing the synergistic mechanism of the Representation Extraction Module and Reasoning Module, along with the collaboration between representation space and sub-enumeration technology, Johnny breaks free from the traditional end-to-end model's direct reliance on option pool configurations. Experimental results demonstrate significant performance improvements across RAVEN, I-RAVEN, and PGM datasets, validating its capability to enable solving models to achieve deep comprehension of problem spaces by supplementing the learned representation space with raw negative configurations.

\item {Spin-Transformer:}  
To address the limitations of Transformer-Encoders in abstract reasoning tasks, this paper proposes the Spin-Transformer architecture. By enhancing the inter-attention head communication mechanism, it strengthens the capture of positional relationships among local features. Experimental results demonstrate that its lightweight variant, Straw Spin-Transformer, achieves comparable performance to the original Spin-Transformer in RPM tasks while significantly reducing computational overhead, offering a viable solution for deployment in resource-constrained scenarios.
 
\item {Theoretical Breakthroughs and Practical Implications :}  
The study highlights the limitations of traditional RPM-solving models' learning paradigms, which exhibit a heavy reliance on observable incorrect options to refine their decision boundaries and reveals the {deep coupling mechanism between representation learning and symbolic logical reasoning}, offering a new pathway for deep learning models to handle abstract concepts like causality and counterfactual reasoning through structured representation spaces.
\end{enumerate}

This study not only provides innovative methodologies for the abstract reasoning field but also lays the technical foundation for the evolution of deep learning models toward advanced cognitive abilities. 
Through continuous optimization of representation learning generalization and computational efficiency, future AI systems are expected to achieve human-like abstract reasoning capabilities across broader domains.

\section{Future Work}

In this section, we will focus on addressing generative RPM problems.
The term ``generative RPM problems" refers to a paradigm where AI algorithms solve RPM problems not by selecting from predefined option pools, but by autonomously generating potential solutions that align with the inherent patterns of the RPM instance.
Generative RPM problems represent an advanced problem, which typically involve regression tasks for solution representation or attribute. Setting aside the inherent complexity of regression tasks themselves, the multi-solution design inherent in RPM problems presents significant challenges for gradient descent-optimized deep learning models. Because gradient descent algorithms are not suited for optimizing and fitting one-to-many mapping functions.
However, the well-trained Johnny ACT3 has initially demonstrated potential in solving generative RPM problems. As demonstrated by Equations (\ref{111}) and (\ref{1111}), Johnny ACT3 exhibits two core capabilities: 
\begin{enumerate}
    \item Identify the most suitable components to serve as solutions for arbitrary RPM problems within the representation space.

    \item Regressing images from components in the representation space that statistically conform to the inherent pixel configuration distribution in RPM images.
\end{enumerate}
In other words, when Johnny ACT3 encounters an RPM instance $\{x_i | i \in [1,8]\}$, its Representation Extraction Module transforms the RPM instance into tokenized representations $\{z_{ij} | i \in [1,8], j \in [1,N]\}$. Thereafter, the Reasoning Module can select $N$ optimal components from the representation spaces $\{T_k | k \in [1,K]\}$ for $\{z_{ij} | i \in [1,8], j \in [1,N]\}$. Furthermore, the decoder $D(\cdot)$ within Johnny ACT3 can regress an image from these $N$ components that statistically conforms to the pixel configuration distribution of RPM images, thereby serving as the solution to the RPM instance  $\{x_i | i \in [1,8]\}$.

This implies that Johnny ACT3 can generate an image it deems optimal as the solution for an RPM instance. If we abandon the optimality principle and instead adopt random sampling based on the probabilistic scores from the reasoning module to select components from the representation space, it could even overcome the multi-solution challenges inherent in RPM problems. However, such a model capable of providing multiple image solutions for RPM instances still faces a fundamental dilemma: evaluating the correctness of the generated solutions. 
The volume of RPM instance used to test AI algorithms' abstract reasoning capabilities is typically substantial. Evaluating Johnny ACT3's generative performance through manual means is impractical.

This paper critiques the end-to-end RPM-solving models' learning approach, arguing that this method overly relies on the comprehensiveness of observed incorrect option configurations. Therefore, using traditional end-to-end RPM-solving models to evaluate Johnny ACT3's performance in addressing generative RPM problems appears inadequate. This paper highlights symbolic RPM-solving models \cite{NVSA}, which require more supervisory signals but offer stronger interpretability, as ideal candidates for evaluating Johnny ACT3's performance. However, given that the current performance of symbolic solving models has not yet matched the level of state-of-the-art approaches \cite{NVSA}, they paradoxically prove less effective than traditional end-to-end models in assessing Johnny ACT3's capabilities. Consequently, this paper proposes that future methodologies for addressing generative RPM problems should in tandem with the symbolic solving model.

\newpage

\end{document}